\documentclass[11pt]{article}

\usepackage[preprint]{acl}

\usepackage{times}
\usepackage{latexsym}

\usepackage[T1]{fontenc}

\usepackage[utf8]{inputenc}

\usepackage{microtype}

\usepackage{inconsolata}

\usepackage{graphicx}
\usepackage{xcolor}
\usepackage{amsmath}
\usepackage{booktabs}
\usepackage{xcolor}
\usepackage{multirow}
\usepackage{array}
\usepackage{colortbl}
\usepackage[table]{xcolor}
\usepackage{booktabs}
\usepackage{adjustbox}

\usepackage{CJKutf8}
\DeclareRobustCommand{\zh}[1]{\begin{CJK*}{UTF8}{gbsn}#1\end{CJK*}}
\usepackage{placeins}

\definecolor{clrEasy}{RGB}{85, 168, 104}
\definecolor{clrMedium}{RGB}{221, 132, 82}
\definecolor{clrHard}{RGB}{196, 78, 82}
\definecolor{clrEntity}{RGB}{76, 114, 176}
\definecolor{clrFactual}{RGB}{196, 78, 82}
\definecolor{clrVersion}{RGB}{221, 132, 82}
\definecolor{clrCriteria}{RGB}{129, 114, 178}
\definecolor{clrSection}{RGB}{230, 235, 248}  
\definecolor{clrRowA}{RGB}{248, 250, 255}
\definecolor{clrRowB}{RGB}{255, 255, 255}

\definecolor{subAcolor}{RGB}{175, 55, 70}
\definecolor{subBcolor}{RGB}{15, 80, 180}
\definecolor{subCcolor}{RGB}{40, 130, 60}
\newcommand{\subA}[1]{\textcolor{subAcolor}{#1}}
\newcommand{\subB}[1]{\textcolor{subBcolor}{#1}}

\definecolor{enBg}{RGB}{244, 244, 245}
\definecolor{zhBg}{RGB}{253, 246, 232}
\definecolor{groupBg}{RGB}{223, 226, 232}

\DeclareUnicodeCharacter{266B}{}

\usepackage{booktabs}
\usepackage[table]{xcolor}
\usepackage{adjustbox}
\usepackage{stfloats}

\definecolor{LightBlue1}{RGB}{235,243,255}
\definecolor{LightBlue2}{RGB}{210,228,252}
\definecolor{LightBlue3}{RGB}{180,210,245}
\definecolor{LightBlue4}{RGB}{140,185,235}
\definecolor{LightBlue5}{RGB}{95,155,225}

\newcommand{\hI}[1]{\cellcolor{LightBlue1}#1}
\newcommand{\hII}[1]{\cellcolor{LightBlue2}#1}
\newcommand{\hIII}[1]{\cellcolor{LightBlue3}#1}
\newcommand{\hIV}[1]{\cellcolor{LightBlue4}#1}
\newcommand{\hV}[1]{\cellcolor{LightBlue5}#1}

\usepackage{listings}
\usepackage{xcolor}
\usepackage[most]{tcolorbox}
\definecolor{promptbg}{gray}{0.97}
\definecolor{promptline}{gray}{0.75}
\usepackage{graphicx}
\lstdefinestyle{aclPromptListing}{
  backgroundcolor=\color{promptbg},
  basicstyle=\ttfamily\scriptsize,
  breaklines=true,
  breakatwhitespace=false,
  columns=fullflexible,
  keepspaces=true,
  showstringspaces=false,
  frame=l,
  rulecolor=\color{promptline},
  framerule=1pt,
  xleftmargin=0.8em,
  xrightmargin=0.2em,
  framexleftmargin=0.5em,
  aboveskip=0.6em,
  belowskip=0.6em,
  captionpos=t,
  numbers=none,
  tabsize=2
}

\title{When Search Agents Should Ask: \textsc{DiscoBench} for Clarification-Aware Deep Search}

\author{
  \textbf{Yiling Tao\textsuperscript{1,2,*,\S}},
  \textbf{Shihan Deng\textsuperscript{1,*}},
  \textbf{Meiling Tao},
\\
  \textbf{Pengzhi Wei\textsuperscript{1}},
  \textbf{Zhichao Hu\textsuperscript{1,\dag}},
  \textbf{Zhihao Zhu\textsuperscript{1,\dag}}
\\
\\
  \textsuperscript{1}Hunyuan, Tencent \qquad
  \textsuperscript{2}Shenzhen International Graduate School, Tsinghua University
}

\begin{document}
\maketitle

\begingroup
\renewcommand{\thefootnote}{\fnsymbol{footnote}}
\footnotetext[1]{Equal contribution.}
\footnotetext[2]{\raggedright Corresponding authors. Correspondence to \href{mailto:elliotzhu@tencent.com}{elliotzhu@tencent.com}\par}
\footnotetext[4]{Work done during an internship at Tencent Hunyuan.}
\endgroup

\begin{abstract}
Search agents powered by large language models (LLMs) are increasingly used to solve complex information-seeking tasks, requiring multi-step retrieval and reasoning to fulfill user goals. However, existing benchmarks often assume that user queries are complete and explicit, overlooking the fact that real-world search requests are frequently vague, underspecified, or even factually incorrect. In deep search scenarios, such ambiguity can propagate along multi-step reasoning chains and lead agents toward incorrect search trajectories. To address this gap, we introduce \textsc{DiscoBench}, a benchmark for clarification-aware deep search, designed to evaluate whether search agents can proactively identify ambiguity, ask effective clarification questions, and recover correct reasoning paths through user interaction. \textsc{DiscoBench} contains 211 samples and 463 ambiguity instances across 11 real-world domains, covering four ambiguity types. We further design a user simulator for multi-turn interaction and evaluate model performance from four perspectives: task utility, ambiguity detection, interaction strategy, and cost efficiency. 
Experiments on representative LLMs show that ambiguity detection and effective clarification are distinct capabilities, and that repeatedly searching instead of asking for clarification often performs worse than direct guessing, highlighting a critical gap between retrieval ability and interactive problem-solving in current search agents.
\end{abstract}

\section{Introduction}
In recent years, search agents based on Large Language Models (LLMs) have made significant progress in the field of information retrieval~\cite{mialon2023gaia,wei2025browsecomp,wong2025widesearch}. The search paradigm is shifting from traditional static corpus retrieval to autonomous Web Search Agents capable of handling complex goals~\cite{xi2025survey,openai2025deepresearch,google2025geminideep}. These agents can simulate human navigation and browsing behaviors, achieving multi-step reasoning and information integration in dynamic and complex internet environments~\cite{wu2025webwalker,zhou2024webarena}. 

\begin{figure}[t]
    \centering
    \includegraphics[width=0.45\textwidth]{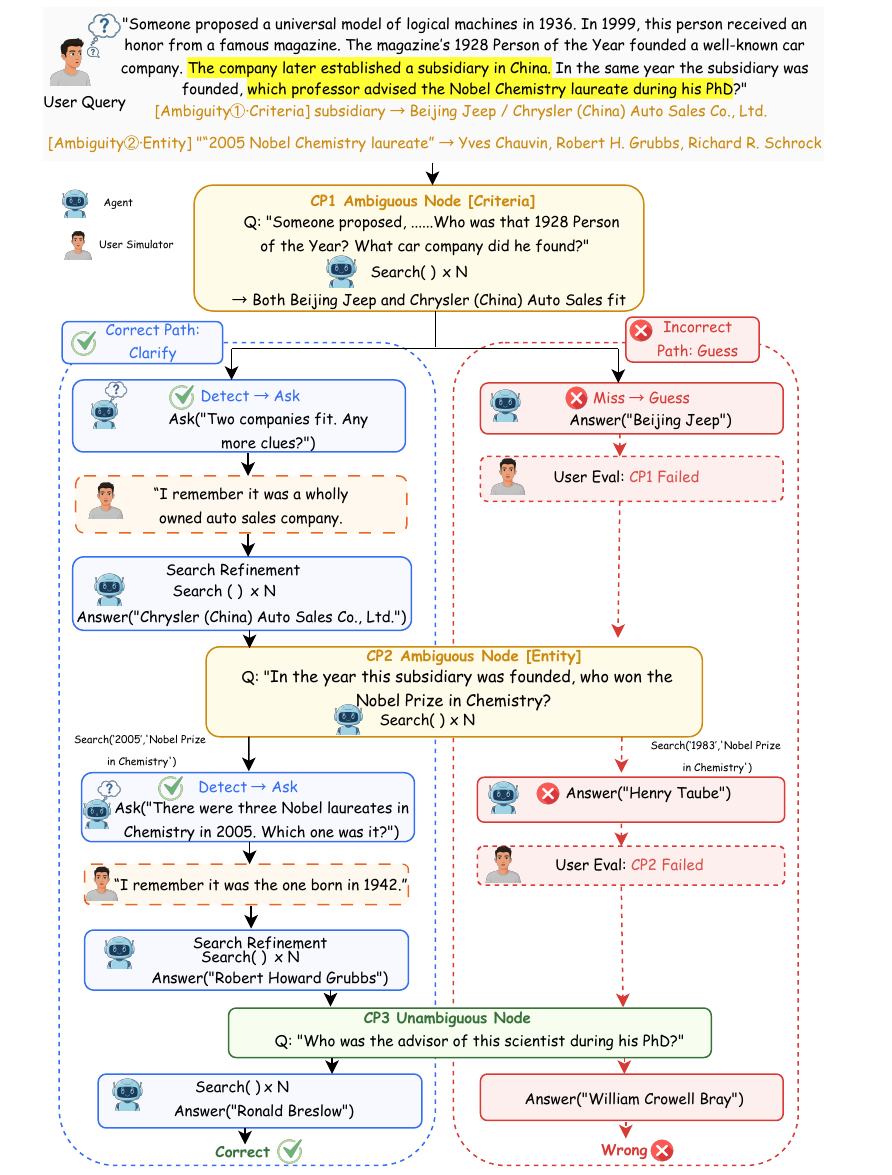}
    \vspace{-1mm}
    \caption{
\small
A motivating example of ambiguity propagation in interactive deep search.
}
    \label{fig:example}
    \vspace{-6mm}
\end{figure}
 

However, current paradigms often presuppose that the user's initial query is complete and explicit. This assumption deviates significantly from real-world information-seeking behavior, where users often provide vague or fragmented queries due to blurred memories or cognitive load limitations~\cite{aliannejadi2019asking,zamani2020analyzing}.
In deep search scenarios, the impact of this discrepancy is further amplified. Unlike traditional single-hop retrieval~\cite{kwiatkowski2019natural}, deep search involves complex multi-step reasoning chains~\cite{trivedi2022musique}, meaning any subtle ambiguity in the initial query can lead to cascading errors in subsequent navigation and information integration, wasting expensive computational resources on the wrong path. As illustrated in Fig.~\ref{fig:example}, failing to proactively clarify ambiguous checkpoints further propagates errors throughout the remaining search process. Consequently, introducing interactive clarification mechanisms to resolve ambiguity has become particularly important. Meanwhile, search tasks naturally provide strong factual grounding, allowing both interaction quality and retrieval correctness to be objectively verified through external evidence. This property offers a reliable signal for evaluating an agent's ability to identify and resolve ambiguity in complex interactive settings.

While the academic community has recognized the importance of query ambiguity and interaction, existing benchmarks still struggle to evaluate the disambiguation capabilities of search agents. Mainstream retrieval benchmarks (e.g., GAIA~\cite{mialon2023gaia}, BrowseComp~\cite{wei2025browsecomp}, AgentBench~\cite{liu2024agentbench}) mostly assume explicit queries and focus on multi-hop reasoning while neglecting proactive interaction. Ambiguity-focused datasets (e.g., AmbigQA~\cite{min2020ambigqa}, DEEPAMBIGQA~\cite{ji2025deepambigqa}) primarily consist of static scenarios and lack dynamic interaction simulation, whereas interaction-based benchmarks (e.g., IN3~\cite{qian2024tell}, UserBench~\cite{qian2025userbench}) are often confined to closed sandbox environments, falling short in the depth and breadth of Web-scale open-domain knowledge. The recent INTERACTCOMP~\cite{deng2025interactcomp} has begun to address search interaction, yet it remains limited in terms of task authenticity, the amplification effects of ambiguity within long-chain reasoning, and the naturalness of interaction modalities.


To bridge these gaps, we introduce \textbf{\textsc{DiscoBench}} (\textbf{D}eep \textbf{I}nteractive \textbf{S}earch with \textbf{C}larificati\textbf{O}n Benchmark)\footnote{Our code and data will be publicly released soon.}, a benchmark for evaluating whether search agents can proactively clarify and resolve ambiguity during multi-step search. Unlike prior benchmarks that mainly focus on static query understanding, \textsc{DiscoBench} models ambiguity as a dynamic phenomenon arising during multi-step search trajectories. 
At each ambiguous checkpoint, agents must proactively identify underspecified information and interact with the user to obtain discriminative clues, rather than relying on direct guessing or closed-form option selection strategies.


We conduct experiments on \textsc{DiscoBench} across a set of representative LLMs. The results show that current search agents still struggle to determine when clarification is needed: even stronger models often fail to recognize ambiguity during the search process or ask effective clarification questions. This suggests that deep interactive search requires not only stronger retrieval and reasoning abilities, but also better ambiguity awareness and proactive clarification strategies. 
Our main contributions are as follows: 
\begin{itemize}
\item We construct \textsc{DiscoBench},  a benchmark that models ambiguity as a dynamic phenomenon propagating along multi-step reasoning chains rather than a static property of individual queries, covering 211 samples with 463 ambiguity instances across 11 real-world domains and four ambiguity types.
\item We propose an ambiguity-aware evaluation framework for multi-turn interactive deep search, together with a user simulator that progressively reveals discriminative clues, enabling unified evaluation of ambiguity detection, clarification effectiveness, and interaction cost. 
\item Through extensive experiments, we reveal that ambiguity detection and effective clarification are distinct capabilities. We further identify a dominant failure mode in which models repeatedly continue searching instead of asking for clarification, leading to lower success rates than direct guessing. This finding highlights the need for mechanisms that explicitly bridge retrieval uncertainty and user interaction.
\end{itemize}

\begin{figure*}[t]
    \centering
    \includegraphics[width=0.9\textwidth]{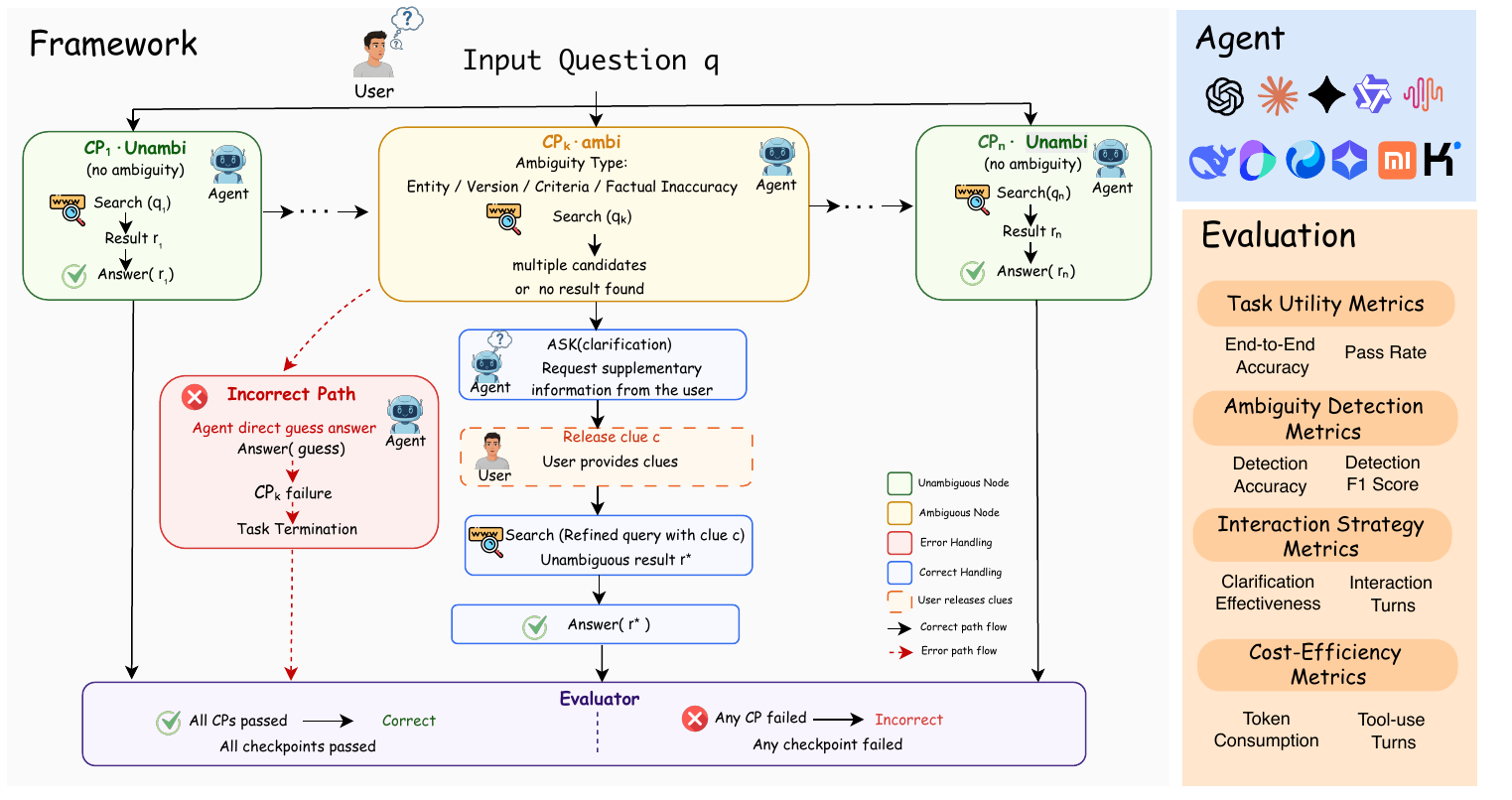}
    \caption{
Overview of the proposed interactive retrieval framework and evaluation protocol.
}
    \label{fig:task_framework}
    \vspace{-4mm}
\end{figure*}

\section{Related Work}
\subsection{Web Search Benchmark}
Efforts to benchmark search agents often bifurcate into two dimensions. One branch focuses on reasoning depth, challenging agents to navigate complex web hierarchies for multi-hop tasks, such as GAIA~\cite{mialon2023gaia} and the BrowseComp series~\cite{wei2025browsecomp,zhou2025browsecomp,chen2025browsecomp}. In parallel, other benchmarks explore information width, necessitating the synthesis of vast horizontal data, as seen in PaSa~\cite{he2025pasa}, SPAR~\cite{shi2025spar}, and WideSearch~\cite{wong2025widesearch}. Complementary agent benchmarks like WebArena~\cite{zhou2024webarena}, VisualWebArena~\cite{koh2024visualwebarena}, Mind2Web~\cite{deng2023mind2web}, and WebShop~\cite{yao2022webshop} further evaluate web navigation capabilities in realistic environments. Recent work, such as DeepWideSearch~\cite{lan2025deepwidesearch}, has further begun to encompass both dimensions. 
Constrained by the assumption of complete queries, these benchmarks leave query ambiguity unexplored, which remains an essential aspect of autonomous search.

\subsection{Ambiguity Benchmark}
Research on query ambiguity has evolved into several taxonomies. Semantic and structural ambiguity benchmarks represented by AmbiEnt~\cite{liu2023we} investigate logical divergences arising from linguistic properties. In the realm of multi-answer factual ambiguity, datasets such as AmbigQA~\cite{min2020ambigqa} and ASQA~\cite{stelmakh2022asqa} focus on mapping single queries to multiple concurrent valid facts. Furthermore, conditional and contextual ambiguity benchmarks including TempAmbigQA~\cite{piryani2024detecting}, CondAmbigQA~\cite{li2025condambigqa} and SituatedQA~\cite{zhang2021situatedqa} address scenarios where answers depend on latent temporal or geographical backgrounds that remain unstated in the initial query. 
While valuable, these benchmarks rely on a static evaluation paradigm that prioritizes answer identification over the dynamic, interactive process required for agent-user collaboration to resolve uncertainty.

\subsection{Interactive Clarification Benchmark}

Interactive benchmarks evaluate agent performance in multi-turn collaborative environments. For instance, ColBench~\cite{zhou2025sweet} and UserBench~\cite{qian2025userbench} target code generation and travel planning, respectively, while IN3~\cite{qian2024tell} and GAIA2~\cite{froger2026gaia2} investigate implicit intent understanding and conflicting requests in local environments. In conversational QA, Abg-CoQA~\cite{guo2021abg} requires agents to clarify coreference or semantic vagueness. Although IDRbench~\cite{feng2026idrbench} recently introduced interactive clarification into deep research, it is not designed for open-domain web search scenarios.
Most closely related, InteractComp~\cite{deng2025interactcomp} pioneers the evaluation of interactive disambiguation for search agents. However, it still contains scenarios that rely more on internal knowledge than external retrieval and mainly focuses on initial entity ambiguity rather than cascading errors in deep reasoning. Moreover, its binary feedback setting cannot fully capture the descriptive nature of real user interactions.

\section{Task Formulation}
We formulate multi-turn interactive retrieval as a sequential question-answering task in which an agent resolves a complex question $q$ through a series of structured checkpoints( $CP$), with the ability to interact with a user when ambiguity is encountered. As illustrated in Fig.~\ref{fig:task_framework}, the agent must determine whether the current retrieval state is ambiguous and decide whether to continue retrieval or request clarification from the user.

\paragraph{Question and Checkpoints.}
Each question $q$ is decomposed into an ordered sequence of $n$ checkpoints $\{CP_1, CP_2, \ldots, CP_n\}$, each representing an intermediate retrieval sub-goal. A checkpoint $CP_i$ is assigned one of two types:
\begin{itemize}
    \item \textbf{Unambi}: an unambiguous checkpoint where the agent can answer directly via retrieval.
    \item \textbf{Ambi}: an ambiguous checkpoint containing one of four injected ambiguity types, which causes retrieval to return multiple candidates or no valid result.
\end{itemize}



\paragraph{Agent Actions and User Interaction.}
At each checkpoint $CP_i$, the agent may execute one of three actions:
\begin{equation}
    a_i \in \{\textsc{Search},\ \textsc{Ask},\ \textsc{Answer}\}
\end{equation}
For unambiguous checkpoints, the agent directly issues $\textsc{Search}(q_i)$ and proceeds with $\textsc{Answer}(r_i)$. For an ambiguous checkpoint $CP_k$, the agent should invoke $\textsc{Ask}(\cdot)$ to request supplementary information, upon which the user releases a pre-defined clue $c$. The agent then refines its search and resolves the ambiguity before issuing $\textsc{Answer}(r^*)$.

\paragraph{Evaluation.}
We evaluate the agent from four aspects: task utility, ambiguity detection, interaction strategy, and cost efficiency. 

\section{Methodology of Dataset Construction}

\begin{figure*}[t]
    \centering
    \includegraphics[width=\textwidth]{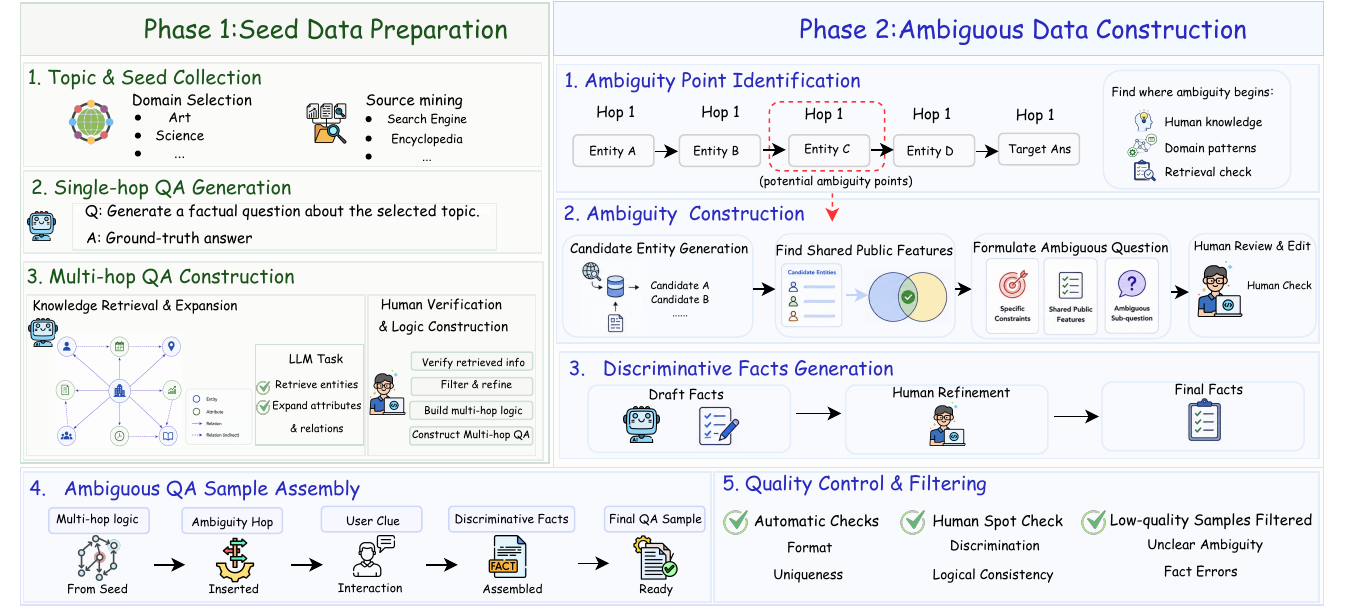}
    \caption{
Overview of the two-phase dataset construction pipeline, including seed multi-hop QA construction, ambiguity injection, discriminative fact generation, and quality control.
    }
    \vspace{-4mm}
    \label{fig:data_construction}
\end{figure*}



We construct \textsc{DiscoBench}, an interactive ambiguous question answering (QA) benchmark designed to evaluate whether LLMs can identify ambiguity, proactively request clarification, and recover correct reasoning trajectories in multi-turn open-domain search tasks.
As illustrated in Fig.~\ref{fig:data_construction}, the construction pipeline consists of two phases: (1) \textit{Seed Data Preparation}, which builds high-quality multi-hop reasoning chains, and (2) \textit{Ambiguous Data Construction}, which injects ambiguity and generates discriminative facts for interactive disambiguation. The entire pipeline adopts a semi-automatic collaborative framework.

\subsection{Seed Data Preparation}

The goal of Phase 1 is to construct high-quality multi-hop seed questions that serve as the foundation for subsequent ambiguity injection.

\paragraph{Topic \& Seed Collection.}

We first manually collect seed topics from 11 diverse knowledge domains to ensure broad domain coverage and knowledge diversity. In terms of knowledge sources, we utilize encyclopedic resources (e.g., Wikipedia and Baidu Baike) together with search engine result pages from search engines (e.g., Google, Bing, and Baidu). \textsc{DiscoBench} is primarily constructed in Chinese to better reflect realistic ambiguity patterns and retrieval behaviors in Chinese web environments. To ensure realistic retrieval requirements, all questions are required to satisfy the following conditions: (1) the answer must be objectively verifiable; (2) the question cannot be solved purely through common sense reasoning; and (3) external retrieval is necessary for task completion.

\paragraph{Multi-hop QA Construction.}
Inspired by existing multi-hop QA datasets~\cite{ho2020constructing,trivedi2022musique}, we adopt a collaborative framework that combines LLM-based preliminary expansion with human verification and reconstruction. Specifically, the LLM first generates preliminary single-hop factual QA pairs based on manually collected seed topics, and further performs graph-structured expansion with external retrieval results to construct candidate multi-hop reasoning chains. 
After automatic generation, human annotators further review and reconstruct the reasoning chains.
Finally, each Seed QA sample is organized into a structured multi-hop instance, serving as the foundation for subsequent ambiguity construction.

\subsection{Ambiguous Data Construction}

Phase 2 aims to inject ambiguity into existing multi-hop reasoning chains, transforming deterministic QA tasks into interactive reasoning tasks that require clarification.

\paragraph{Ambiguity Point Identification.}

Given a deterministic multi-hop reasoning chain, we identify hops where ambiguity can be naturally introduced. Instead of injecting ambiguity randomly, we focus on nodes whose target entity has similar alternatives, such that relaxing the distinguishing constraint leads to multiple plausible candidates. A node is retained as a candidate ambiguity point if: (1) its target entity shares attributes with sibling entities; (2) the downstream reasoning chain remains executable under underspecification; and (3) the ambiguity can be resolved with a single user-provided clue. All candidate positions are further verified manually.

\paragraph{Ambiguity Construction.}

After identifying ambiguity points, we inject ambiguity into the original reasoning chain by replacing strong constraints with shared attributes among multiple candidate entities. Specifically, the system retrieves candidate entities satisfying the current reasoning constraints and uses LLMs to identify shared characteristics, such as common authors, temporal ranges, or organizational relations. The original question is then rewritten using these shared features, allowing multiple candidates to satisfy the same description. Human annotators further verify the naturalness, solvability, and logical consistency of the constructed ambiguous questions.



\paragraph{Discriminative Facts Generation.}

To support interactive disambiguation, we construct discriminative facts for each ambiguity point, which simulate supplementary clues provided by users and distinguish the target entity from distractors. Candidate facts are generated by retrieval-augmented LLMs from perspectives such as entity attributes, temporal information, relations, numerical facts, version differences, and organizational associations, and are then manually verified for factual correctness, distinguishability, and naturalness.






\subsection{Data Statistics and Quality Control}
\label{sec:data_quality}

Tab.~\ref{tab:data_statistics} presents the overall statistics of \textsc{DiscoBench}, including domain distribution, task difficulty, and ambiguity types. Task difficulty is determined by the number of ambiguity checkpoints, where easy, medium, and hard correspond to 1, 2, and 3 ambiguity points, respectively. \textsc{DiscoBench} covers four ambiguity types: \textit{Entity} (multiple entities satisfy the same description), \textit{Version} (different temporal or version-specific states), \textit{Criteria} (different evaluation standards or ranking criteria), and \textit{Factual Inaccuracy} (descriptions inconsistent with objective facts).

\textsc{DiscoBench} construction process involved an expert annotation instructor, six undergraduate annotators, and two quality inspectors. Annotators are recruited from diverse academic backgrounds in multiple disciplines to ensure annotation diversity and broad domain coverage. During the construction process, all samples were further reviewed for factual correctness, retrieval feasibility, logical consistency, and ambiguity solvability.
\begin{table}[t]
\centering
\scriptsize
\setlength{\tabcolsep}{1.8pt}
\renewcommand{\arraystretch}{0.98}

\caption{\small Data statistics of \textsc{DiscoBench} (211 samples, 463 ambiguity instances).}
\label{tab:data_statistics}

\begin{tabular}{@{}lrr@{\hspace{2pt}}|@{\hspace{2pt}}lrr@{\hspace{2pt}}|@{\hspace{2pt}}lrr@{\hspace{2pt}}lrr@{}}
\toprule

\multicolumn{3}{c}{\textbf{Difficulty}} &
\multicolumn{3}{c}{\textbf{Ambiguity}} &
\multicolumn{6}{c}{\textbf{Domain}} \\

\cmidrule(r{2pt}){1-3}
\cmidrule(r{2pt}){4-6}
\cmidrule(l{2pt}){7-12}

\textbf{Level} & \textbf{n} & \textbf{\%} &
\textbf{Type} & \textbf{n} & \textbf{\%} &
\textbf{Type} & \textbf{n} & \textbf{\%} & 
\textbf{Type} & \textbf{n} & \textbf{\%} \\

\midrule

\rowcolor{clrRowA}
Easy & 44 & 20.9 &
Entity & 176 & 38.0 &
Film/TV & 32 & 15.2 &
Sports & 18 & 8.5 \\

\rowcolor{clrRowB}
Medium & 82 & 38.9 &
Factual & 125 & 27.0 &
Games & 27 & 12.8 &
Geography & 18 & 8.5 \\

\rowcolor{clrRowA}
Hard & 85 & 40.3 &
Version & 109 & 23.5 &
Academic & 22 & 10.4 &
Music & 16 & 7.6 \\

\rowcolor{clrRowB}
& & &
Criteria & 53 & 11.4 &
Art & 19 & 9.0 &
Medicine & 16 & 7.6 \\

\rowcolor{clrRowA}
& & &
& & &
Finance & 18 & 8.5 &
Tech. & 15 & 7.1 \\

\rowcolor{clrRowB}
& & &
& & &
& & &
Policy/Law & 10 & 4.7 \\

\bottomrule
\vspace{-4mm}
\end{tabular}

\end{table}

\section{Experiments}
\subsection{Experimental Setup}

\begin{table*}[t]
    \centering
    \small
    \caption{
\small
Main results on \textsc{DiscoBench} under Neutral/Guided prompting.
Acc.: end-to-end accuracy;
CP: checkpoint pass rate;
Det.: ambiguity detection;
CE: clarification evaluation;
Ask: average clarification turns.
Darker blue indicates stronger neutral-prompt performance.
}
    \label{tab:main_results}
 \begin{tabular}{lccccccc}
        \toprule
        \textbf{Model} 
        & \textbf{Acc.(\%)} 
        & \textbf{CP(\%)} 
        & \textbf{Det. Acc.(\%)} 
        & \textbf{Det. F1(\%)} 
        & \textbf{CE-A(\%)} 
        & \textbf{CE-B(\%)} 
        & \textbf{Ask} \\
        \midrule

        Doubao-Seed-2.0-Pro
        & \hV{\textbf{43.1}/50.2} 
        & \hV{\textbf{63.6}/70.4} 
        & \hV{68.6/75.4} 
        & \hV{61.9/73.9} 
        & \hV{93.8/87.5} 
        & \hV{89.2/83.3} 
        & 0.84/1.40 \\

        Gemini-3.1-Pro-Preview 
        & \hV{40.8/\textbf{53.1}} 
        & \hV{62.1/\textbf{73.8}} 
        & \hV{\textbf{69.8}/\textbf{75.7}} 
        & \hV{\textbf{64.5}/\textbf{75.6}} 
        & \hI{87.8/88.3} 
        & \hII{82.2/83.2} 
        & 0.81/1.40 \\

        Claude-Opus-4.7 
        & \hIV{39.8/38.9} 
        & \hIV{57.0/61.6} 
        & \hIV{60.7/71.3} 
        & \hIV{48.9/68.9} 
        & \hIV{92.0/90.0} 
        & \hIV{88.3/82.6} 
        & 0.58/1.05 \\

        DeepSeek-V4-Pro 
        & \hIV{35.5/38.9} 
        & \hIV{57.6/62.2} 
        & \hIV{60.8/68.8} 
        & \hIV{48.6/63.7} 
        & \hII{87.9/\textbf{92.9}} 
        & \hIII{82.3/\textbf{85.8}} 
        & 0.64/1.10 \\

        Kimi-K2.6 
        & \hIII{29.4/35.1} 
        & \hIII{51.3/61.3} 
        & \hII{57.0/73.3} 
        & \hII{42.4/71.0} 
        & \hII{90.1/90.5} 
        & \hIII{83.8/83.4} 
        & 0.54/1.14 \\

        GLM-5.1 
        & \hIII{28.4/38.4} 
        & \hII{50.1/61.6} 
        & \hII{57.9/72.1} 
        & \hII{44.2/69.8} 
        & \hIV{91.7/92.3} 
        & \hIV{86.2/85.3} 
        & 0.49/1.19 \\

        GPT-5.4$^{\dagger}$ 
        & \hII{27.5/-} 
        & \hIII{54.9/-} 
        & \hIII{58.9/-} 
        & \hIII{45.1/-} 
        & \hI{87.5/-} 
        & \hI{76.0/-} 
        & 0.51/- \\

        MiMo-v2.5-Pro 
        & \hII{24.2/28.4} 
        & \hII{46.3/55.8} 
        & \hIII{58.5/65.0} 
        & \hIII{47.2/59.5} 
        & \hIII{91.1/85.9} 
        & 75.0/76.2 
        & 0.60/1.04 \\

        Hunyuan-3.0-Preview  
        & \hI{16.1/24.2} 
        & \hI{40.4/47.3} 
        & \hI{55.9/67.6} 
        & \hI{40.0/62.1} 
        & \hIII{90.2/92.1} 
        & \hII{78.3/82.5} 
        & 0.34/0.70 \\

        MiniMax-M2.7 
        & 16.1/15.2 
        & 39.0/42.4 
        & 53.3/59.3 
        & 39.6/52.4 
        & 78.6/83.5 
        & 60.7/66.5 
        & 0.61/1.10 \\

        Qwen3.6-Max 
        & 12.3/14.6 
        & 33.1/39.9 
        & 51.5/63.4 
        & 16.0/51.8 
        & \hV{\textbf{94.7}/90.8} 
        & \hV{\textbf{89.5}/85.1} 
        & 0.07/0.42 \\

        \bottomrule

    \end{tabular}
    \vspace{-1mm}
    
{\tiny $^{\dagger}$ GPT-5.4 failed on 37 neutral-prompt questions due to usage-policy filtering; guided results are omitted due to only 62 valid runs, so this model is excluded from subsequent analysis.}
    \vspace{-4mm}
\end{table*}


\paragraph{Models and Tools.}
We evaluate Claude-Opus-4.7, GPT-5.4, Gemini-3.1-Pro-Preview, Doubao-Seed-2.0-Pro-High, DeepSeek-V4-Pro, Qwen-3.6-Max, MiniMax-M2.7, GLM-5.1, MiMo-v2.5-Pro, Kimi-K2.6, and Hunyuan-3.0-Preview under the same interactive retrieval framework and checkpoint-level evaluator. For models supporting configurable reasoning effort, we use the maximum available reasoning-effort setting in the main experiments. All \textsc{Search} calls are implemented using Tavily~\cite{tavily2026} as the backend search engine. We use Gemini-3-Flash-Medium as the simulated user model for multi-turn interaction and ambiguity clarification during evaluation.

\paragraph{Prompting Settings.}
We consider two prompting settings. In the Neutral setting, the agent receives no explicit instruction that ambiguity may exist and must independently decide whether clarification is needed. This setting evaluates the model's spontaneous ambiguity detection and proactive interaction ability. In the Guided setting, the prompt explicitly reminds the agent to be aware of potential ambiguity and to ask clarification questions when necessary, which provides an ambiguity-aware condition and reflects the model's upper-bound performance when it is encouraged to interact.

\paragraph{Metrics.}
We report metrics from four aspects. For task utility, we use end-to-end accuracy and checkpoint pass rate. For ambiguity detection, we report detection accuracy and detection F1. For interaction quality, we report the accuracy of the clarification question (CE-A) and the clarification-to-advance rate (CE-B). For cost efficiency, we report average ask turns, tool-use turns, and token consumption. Detailed definitions of all metrics and additional analysis of token consumption are provided in Appendix~\ref{app:metrics} and Appendix~\ref{app:token_consumption}.

\subsection{Main Results}
\paragraph{Frontier models still struggle with clarification-aware deep search.}
As shown in Tab.~\ref{tab:main_results}, current frontier models still show limited performance on \textsc{DiscoBench}. Under the Neutral setting, the best-performing model, Doubao-Seed-2.0-Pro, achieves only 43.1\% end-to-end accuracy, while Gemini-3.1-Pro reaches 40.8\%. Most other models remain below 40\%, and weaker models such as MiniMax-M2.7 and Qwen3.6-Max achieve only 16.1\% and 12.3\%, respectively. At the same time, there is still a substantial gap between the pass rate of the checkpoint and end-to-end accuracy. For example, Claude-Opus-4.7 achieves a checkpoint pass rate of 57.0\% but only 39.8\% accuracy. This suggests that models may solve several intermediate retrieval steps while still failing to complete the full reasoning trajectory due to unresolved ambiguity. Therefore, deep search for clarity requires not only retrieval and reasoning ability, but also stable ambiguity recognition and interaction planning throughout the reasoning process.

\paragraph{Guided prompting improves performance but still reveals limited clarification ability.}
Guided prompting generally improves model performance by explicitly encouraging the agent to identify ambiguity and ask clarification questions when necessary. Averaged over the 10 models with valid results under both settings, end-to-end accuracy increases from 28.6\% to 33.7\%, checkpoint pass rate rises from 50.1\% to 57.6\%, and detection F1 improves substantially from 45.3\% to 64.9\%. The improvement is mainly reflected in ambiguity detection rather than downstream reasoning, suggesting that Guided prompting primarily helps reduce missed ambiguity cases. However, additional interaction does not always translate into better end-to-end performance. For example, Claude-Opus-4.7 achieves a higher checkpoint pass rate under Guided prompting while slightly decreasing in final accuracy, indicating that stronger local interaction behavior may still fail to recover the complete reasoning trajectory. Overall, prompt engineering can partially activate ambiguity-aware behavior, but current models still lack robust and stable clarification ability. Additional analysis by reasoning effort is provided in Appendix~\ref{app:reasoning_effort}, showing that higher reasoning effort improves performance.

\begin{figure}[t]
    \centering
    \input{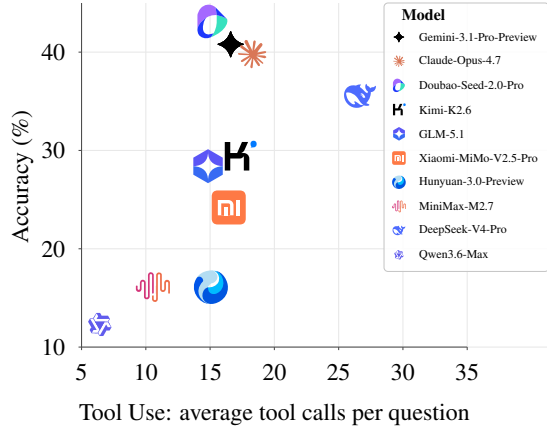}
    \caption{
    \small 
    Performance--efficiency trade-off under neutral prompting. 
    }
    \label{fig:efficiency_tradeoff}
    \vspace{-4mm}
\end{figure}

\paragraph{Knowing when to ask and asking effectively are distinct capabilities.}
Detection and clarification metrics capture different aspects of proactive interaction: recognizing when clarification is needed and asking questions that effectively resolve ambiguity. These abilities are not always aligned. Qwen3.6-Max has only 16.0\% detection F1 and asks just 0.07 questions per task under the Neutral setting, but achieves 94.7\% CE-A and 89.5\% CE-B, indicating strong conditional question quality but weak proactive clarification. By contrast, MiniMax-M2.7 asks more often, with 0.61/1.10 asks under Neutral/Guided settings, yet its CE-B remains lower at 60.7\%/66.5\%. Thus, successful clarification-aware search requires both ambiguity detection and effective question asking.

\paragraph{More tool use does not necessarily lead to better performance.}
Fig.~\ref{fig:efficiency_tradeoff} further reveals that higher retrieval intensity does not consistently translate into better task performance. Increasing search tool calls alone cannot reliably improve accuracy. For example, Claude-Opus-4.7 exhibits relatively high tool-use frequency among evaluated models, yet its accuracy still remains below Gemini-3.1-Pro and Doubao-Seed-2.0-Pro. Meanwhile, several models also perform frequent retrieval actions while still achieving poor end-to-end performance. These observations suggest that \textsc{DiscoBench} does not reward excessive or inefficient retrieval behavior. Successful clarification-aware deep search depends not on searching more, but on whether models can strategically allocate retrieval actions, identify ambiguity at the correct checkpoints, and effectively utilize retrieved evidence and user-provided clues to recover the reasoning trajectory.

\subsection{Performance by Ambiguity Types}

\begin{figure*}[t]
    \centering
    \includegraphics[width=0.95\textwidth]{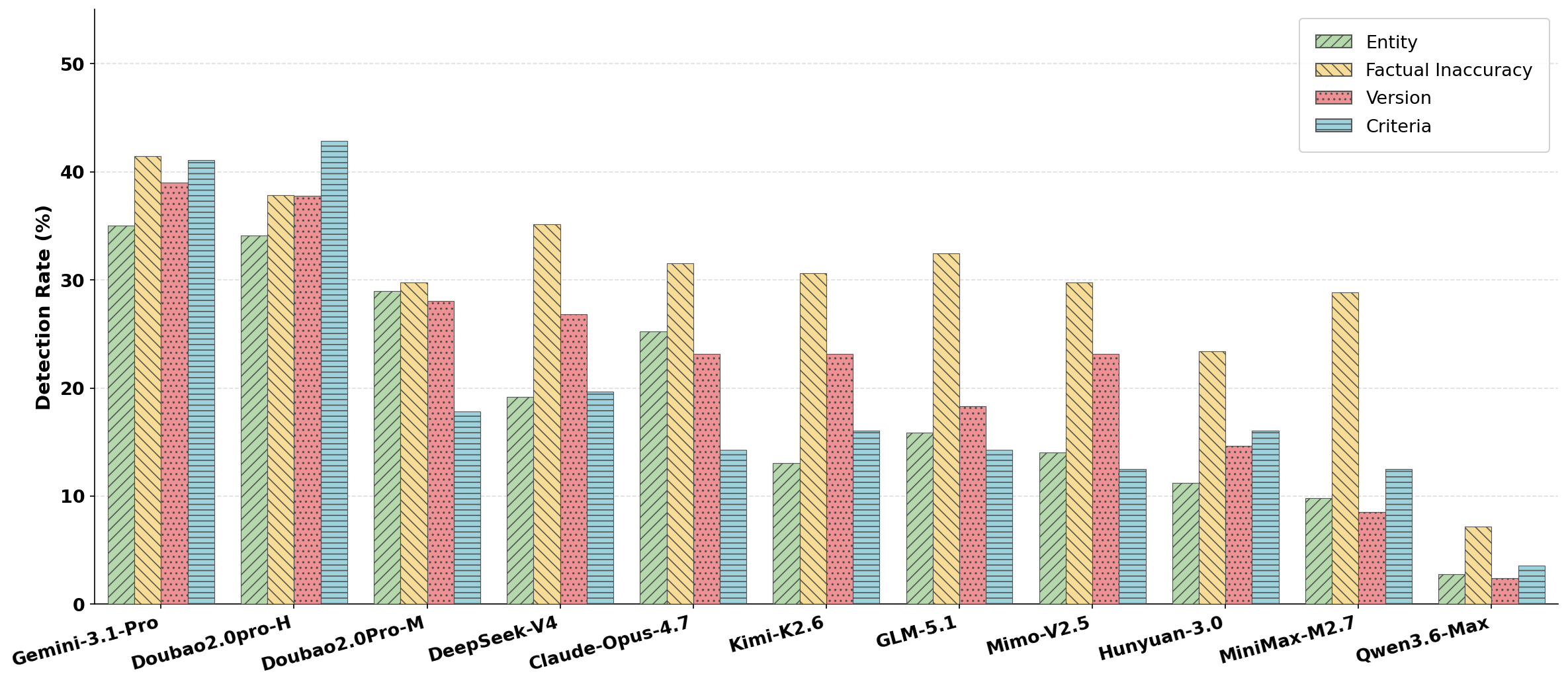}
    \vspace{-2mm}
    \caption{
    Detection performance across different ambiguity types. }
    \label{fig:ambiguity_complexity}
    \vspace{-4mm}
\end{figure*}

Fig.~\ref{fig:ambiguity_complexity} shows that models exhibit clear differences in detection performance across ambiguity types. Stronger models, such as Gemini-3.1-Pro and Doubao-Seed-2.0-Pro, generally achieve more balanced performance, while mid-performing models, such as DeepSeek-V4-Pro and Claude-Opus-4.7, show a more type-dependent pattern. In particular, \textit{Factual Inaccuracy} is often easier to detect, likely because factual errors tend to create explicit conflicts with retrieved evidence, helping models recognize that the current question cannot be directly resolved.
In contrast, \textit{Entity} and \textit{Criteria} ambiguities are more challenging because they usually do not create explicit factual conflicts. Instead, they require models to distinguish among multiple plausible entities or identify missing decision criteria, making models more likely to follow one plausible path prematurely. This suggests that current search agents still struggle with ambiguities that require active clarification rather than direct fact checking. Additional analysis by ambiguity complexity is provided in Appendix~\ref{app:complexity}.

\subsection{Behavioral Profile Analysis}
\label{sec:profile_analysis}

\begin{table}[t]
\centering
\caption{
\small
Pass rate by behavioral profile on the common subset ($N=146$ ambi-CPs).
}
\label{tab:profile_pass}

\footnotesize
\setlength{\tabcolsep}{5pt}
\renewcommand{\arraystretch}{1.05}

\begin{tabular}{lccc}
\toprule
\textbf{Model} & \textbf{DG(\%)} & \textbf{SHG(\%)} & \textbf{STA(\%)} \\
\midrule
Gemini-3.1-Pro   & 66.7 & 56.5$^*$ & \textbf{96.2} \\
Doubao-Seed-2.0-Pro   & 57.1$^*$ & 55.3 & \textbf{98.6} \\
DeepSeek-V4-Pro & 63.2 & 50.8 & \textbf{95.7} \\
Claude-Opus-4.7   & 58.3 & 55.0 & \textbf{97.8} \\
MiMo-v2.5-Pro     & 53.6$^*$ & 53.3 & \textbf{83.7} \\
GLM-5.1      & 53.3 & 48.9 & \textbf{92.7} \\
MiniMax-M2.7  & 50.0 & 51.5 & \textbf{91.2} \\
Kimi-K2.6   & 54.9 & 52.5 & \textbf{91.2} \\
Hunyuan-3.0-Preview  & 51.4 & 42.9 & \textbf{93.9} \\
\midrule
\textbf{Mean} & 56.5 & 51.9 & \textbf{93.4} \\
\bottomrule
\end{tabular}

{\scriptsize
DG denotes DirectGuess; 
SHG denotes SearchHeavyGuess; 
STA denotes SearchThenAsk; 
$^*$ denotes $N<30$.
}
\vspace{-4mm}

\end{table}
To better understand the behavioral differences behind model performance, we categorize ambiguous-checkpoint trajectories into four interaction profiles based on the ordering of \textsc{Search} and \textsc{Ask} actions: DirectGuess, SearchHeavyGuess, DirectAsk, and SearchThenAsk. Detailed definitions and profile distributions are provided in Appendix~\ref{app:profile_details}.

\paragraph{Clarification substantially improves success rates.}
As shown in Tab.~\ref{tab:profile_pass}, SearchThenAsk consistently achieves the highest pass rate across all evaluated models, reaching an average of 93.4\%, substantially outperforming DirectGuess (56.5\%) and SearchHeavyGuess (51.9\%). The gap remains stable within every model, indicating that proactive clarification is critical for successful ambiguity resolution in deep search.


\paragraph{Search-heavy guessing reveals a major failure mode.}
Notably, SearchHeavyGuess even underperforms DirectGuess despite performing more retrieval steps. Repeated retrieval often indicates that the model is already aware of multiple candidate entities. Therefore, these failures arise not from completely missing ambiguity, but from failing to escalate retrieval uncertainty into clarification. This finding further explains why increased tool use alone does not reliably improve performance.

\subsection{Ablation Study}

\paragraph{Effect of Search Tool.}
Tab.~\ref{tab:ablation} shows that the external search tool is crucial for \textsc{DiscoBench}. After removing the search tool, all models suffer substantial accuracy drops. For example, Doubao-Seed-2.0-Pro decreases from 43.1\% to 2.4\%, with a drop of 40.7 points; Gemini-3.1-Pro and DeepSeek-V4-Pro also drop by 20.9 and 25.7 points, respectively. This indicates that \textsc{DiscoBench} cannot be solved by relying solely on parametric knowledge. Models need external retrieval to gather evidence, verify intermediate constraints, and continuously revise the search trajectory.

\paragraph{Effect of Ambiguity.}
The comparison with unambiguous questions further shows that ambiguity is a major source of task difficulty. After removing ambiguity, all models achieve significantly higher accuracy, with improvements ranging from 26.8\% to 40.2\%. This suggests that current search agents are better at answering well-specified questions, but still easily fail when facing ambiguous ones.

\begin{table}[t]
    \centering
     \caption{
     \small
    Ablation results under neutral prompting.
    Full denotes the original \textsc{DiscoBench} accuracy from the main setting.
    }
    \footnotesize
    \setlength{\tabcolsep}{2.2pt}
    \begin{adjustbox}{max width=\columnwidth}
    \begin{tabular}{lccccc}
        \toprule
        \multirow{2}{*}{\textbf{Model}} 
        & \multirow{2}{*}{\textbf{Full(\%)}} 
        & \multicolumn{2}{c}{\textbf{w/o Search}} 
        & \multicolumn{2}{c}{\textbf{Unambig. Qs.}} \\
        \cmidrule(lr){3-4}\cmidrule(lr){5-6}
        & 
        & \textbf{Acc.(\%)} 
        & \textbf{$\Delta$} 
        & \textbf{Acc.(\%)} 
        & \textbf{$\Delta$} \\
        \midrule
        Doubao-Seed-2.0-Pro& 43.1 & 2.4 & -40.7 & 71.4 & +28.3  \\
        Gemini-3.1-Pro            & 40.8 & 19.9 & -20.9 & 81.0 & +40.2 \\
        DeepSeek-V4-Pro           & 35.5 & 9.8 & -25.7 & 74.4 & +38.9 \\
        Hunyuan-3.0-Preview       & 16.1 & 2.9 & -13.2 & 45.5 & +29.4 \\
        MiniMax-M2.7              & 16.1 & 0.8 & -15.3 & 42.9 & +26.8 \\
        \bottomrule
    \end{tabular}
    \end{adjustbox}
    \label{tab:ablation}
    \vspace{-4mm}
\end{table}

\section{Conclusion}
We introduced \textsc{DiscoBench}, a benchmark for evaluating clarification-aware deep search. \textsc{DiscoBench} models ambiguity as a dynamic issue that emerges during multi-step search and uses structured checkpoints to evaluate whether search agents can detect ambiguity, ask for clarification, and recover correct reasoning paths with user-provided clues. Experiments show that current LLM-based search agents still struggle with interactive deep search. Guided prompting improves ambiguity detection, but end-to-end performance remains limited. Meanwhile, proactive clarification is substantially more effective than repeated search or direct guessing. These findings suggest that future search agents need not only stronger retrieval and reasoning abilities, but also better ambiguity awareness and interaction planning.

\section*{Limitations}

\textsc{DiscoBench} primarily focuses on four representative ambiguity types grounded in objective question answering scenarios. More complex forms of ambiguity, such as subjective preference ambiguity, remain underexplored and are left for future work.
In addition, although \textsc{DiscoBench} employs an ambiguity-aware multi-turn user simulator with progressive clue disclosure, the interaction behavior is still generated by LLMs rather than real human users. As a result, the current simulator may not fully capture the diversity, noisiness, and unpredictability of real-world clarification interactions.

\section*{Ethical Considerations}
\textsc{DiscoBench} is constructed from publicly available web resources, including encyclopedic websites and search engine result pages, and does not involve private or personally identifiable information. The benchmark is designed solely for research purposes to evaluate ambiguity handling and clarification abilities in search agents. Although the user simulator is LLM-based rather than collected from real users, we acknowledge that simulated interactions may not fully reflect the diversity of real-world human behavior.


\bibliography{reference}

\appendix

\section{Author Contributions}
\paragraph{Benchmark Design and Methodology.}

Yiling Tao, Shihan Deng, Zhihao Zhu, and Pengzhi Wei jointly contributed to the design of \textsc{DiscoBench}, and the overall methodology was integrated and proposed by Shihan Deng and Yiling Tao.

\paragraph{Data Construction and Annotation.}

Shihan Deng led the construction of the multi-hop seed data, while Yiling Tao led the ambiguity data construction pipeline. Shihan Deng, Zhihao Zhu, Yiling Tao, and Pengzhi Wei were responsible for the critical quality control and final verification of the constructed data.

\paragraph{Evaluation Framework and User Simulator.}

Yiling Tao and Shihan Deng jointly designed the framework, and Shihan Deng was primarily responsible for its code construction and implementation.

\paragraph{Experiments and Analysis.}

Shihan Deng conducted the main experiments. The result analysis was performed by Yiling Tao, Shihan Deng, and Zhihao Zhu.

\paragraph{Paper Writing.}

Yiling Tao led the paper writing, and Meiling Tao produced all the figures and organized the key information. Zhihao Zhu and Shihan Deng participated in revising the manuscript.

\paragraph{Project Supervision.}

Zhichao Hu and Zhihao Zhu supervised the project and served as the corresponding authors.

\section{Evaluation Metrics}
\label{app:metrics}

This section provides detailed definitions of the evaluation metrics used in our experiments. 
All metrics are computed at the question level or checkpoint level and then averaged over all valid questions for each model.

\subsection{End-to-End Accuracy}
\label{sec:appendix-e2e}

End-to-end accuracy evaluates whether the agent's final answer to the
full question matches the ground-truth answer. Equivalence between
the two answers is determined by an LLM-based answer-equivalence
judge, which abstracts away surface-form
variations such as transliterations, date formats, and list ordering.

For each question $q$, let $\hat{a}_q$ denote the agent's final
answer and $a_q^{\star}$ the ground-truth answer, and let
$\mathrm{equiv}(\hat{a}_q, a_q^{\star}) \in \{0, 1\}$ denote the
judge's binary verdict, where $1$ indicates that the two answers are
judged equivalent and $0$ otherwise. The per-question correctness
indicator is defined as:
\begin{equation}
\mathrm{Acc}(q) = \mathrm{equiv}\!\left(\hat{a}_q,\, a_q^{\star}\right).
\end{equation}

The model-level end-to-end accuracy is computed by averaging over all
valid questions:
\begin{equation}
\mathrm{Acc} = \frac{1}{|Q|}\sum_{q \in Q} \mathrm{Acc}(q),
\end{equation}
where $Q$ denotes the set of valid evaluated questions. This
question-level normalization ensures that each question contributes
equally to the final score.

\subsection{Checkpoint Pass Rate}
\label{app:cp_rate}

Each question in \textsc{DiscoBench} is decomposed into a sequence of checkpoints. 
A checkpoint is counted as successfully advanced if the agent either answers the checkpoint correctly and proceeds to the next checkpoint, or correctly completes the final checkpoint. 
In our evaluator logs, this corresponds to one of two outcomes: 
(1) \texttt{correct\_answer}, where the agent correctly resolves a regular checkpoint; 
(2) \texttt{missed\_ambiguity\_correct}, where the agent misses the ambiguity at an ambiguous checkpoint but still happens to answer it correctly.

For each question $q$, let $N_q$ denote the ground-truth number of checkpoints, and let $A_q$ denote the number of checkpoints that are successfully advanced. 
The checkpoint pass score for question $q$ is defined as:
\begin{equation}
    \text{CP}(q) = \frac{A_q}{N_q}.
\end{equation}

The model-level checkpoint pass rate is computed by averaging over all valid questions:
\begin{equation}
    \text{CP} = \frac{1}{|\mathcal{Q}|}\sum_{q \in \mathcal{Q}} \text{CP}(q),
\end{equation}
where $\mathcal{Q}$ denotes the set of valid evaluated questions. 
This question-level normalization ensures that each question contributes equally to the final score, regardless of how many checkpoints it contains.

\subsection{Ambiguity Detection Metrics}
\label{app:detection}

We evaluate ambiguity detection at the checkpoint level. 
For each reached checkpoint, we compare the ground-truth checkpoint type with the agent's interaction behavior. 
A checkpoint is labeled as \textit{Ambi} if it contains an injected ambiguity, and \textit{Non-Ambi} otherwise. 
We define the four detection outcomes as follows:
\begin{itemize}
    \item \textbf{True Positive (TP)}: the checkpoint is \textit{Ambi}, and the agent correctly asks a clarification question targeting the ambiguity.
    \item \textbf{False Negative (FN)}: the checkpoint is \textit{Ambi}, but the agent fails to ask or does not correctly target the ambiguity.
    \item \textbf{False Positive (FP)}: the checkpoint is \textit{Non-Ambi}, but the agent unnecessarily asks for clarification.
    \item \textbf{True Negative (TN)}: the checkpoint is \textit{Non-Ambi}, and the agent proceeds without asking for clarification.
\end{itemize}

Accordingly, $TP+FN$ corresponds to all reached ambiguous checkpoints, while $FP+TN$ corresponds to all reached non-ambiguous checkpoints. 
The total number of evaluated detection decisions is $TP+TN+FP+FN$.

\paragraph{Detection Accuracy.}
Detection accuracy measures the overall correctness of ambiguity detection decisions:
\begin{equation}
    \text{Detection Accuracy} = 
    \frac{TP + TN}{TP + TN + FP + FN}.
\end{equation}

\paragraph{Detection F1 Score.}
We further report detection F1 to better account for the imbalance between ambiguous and non-ambiguous checkpoints. 
Precision and recall are defined as:
\begin{equation}
    P = \frac{TP}{TP + FP}, 
    \quad
    R = \frac{TP}{TP + FN}.
\end{equation}

The detection F1 score is computed as:
\begin{equation}
    \text{Detection F1} = 
    \frac{2 \cdot P \cdot R}{P + R}.
\end{equation}

Here, recall measures the proportion of ambiguous checkpoints that are correctly detected, while precision measures how often the agent's clarification decisions are correct. 
F1 provides a more robust measure when ambiguous and non-ambiguous checkpoints are unevenly distributed.

\subsection{Clarification Effectiveness}
\label{app:clarification}

Clarification effectiveness evaluates the quality and usefulness of the agent's clarification behavior. 
Both metrics share the same denominator: the number of checkpoints where the agent actively invokes \textsc{Ask}, regardless of whether the question is correct.

Let $\mathcal{C}$ denote the set of evaluated checkpoints, and let $\mathcal{C}_{\texttt{asked}}$ denote the set of checkpoints where the agent invokes \textsc{Ask}:
\begin{equation}
    \mathcal{C}_{\texttt{asked}} =
    \{c \in \mathcal{C}: \texttt{asked}(c)\}.
\end{equation}

We further define:
\begin{equation}
    \mathcal{C}_{\texttt{right}} =
    \{c \in \mathcal{C}_{\texttt{asked}} : \texttt{asked\_right}(c)\},
\end{equation}
\begin{equation}
    \mathcal{C}_{\texttt{adv}} =
    \{c \in \mathcal{C}_{\texttt{asked}} : \texttt{cp\_advanced}(c)\}.
\end{equation}

\paragraph{CE-A: Clarification Question Accuracy.}
CE-A measures whether the agent asks the right clarification question when it decides to interact:
\begin{equation}
    \text{CE-A} =
    \frac{
    \left|\mathcal{C}_{\texttt{right}}\right|
    }{
    \left|\mathcal{C}_{\texttt{asked}}\right|
    }.
\end{equation}

\paragraph{CE-B: Clarification-to-Advance Rate.}
CE-B measures whether a correct clarification eventually helps the agent advance the current checkpoint:
\begin{equation}
    \text{CE-B} =
    \frac{
    \left|\mathcal{C}_{\texttt{right}} \cap \mathcal{C}_{\texttt{adv}}\right|
    }{
    \left|\mathcal{C}_{\texttt{asked}}\right|
    }.
\end{equation}

CE-A reflects whether the agent asks in the correct direction, while CE-B further evaluates whether the agent can use the returned clue to successfully advance the checkpoint.

\section{Additional Analysis by Reasoning Effort}
\label{app:reasoning_effort}
\begin{figure}[t]
    \centering
    \includegraphics[width=\columnwidth]{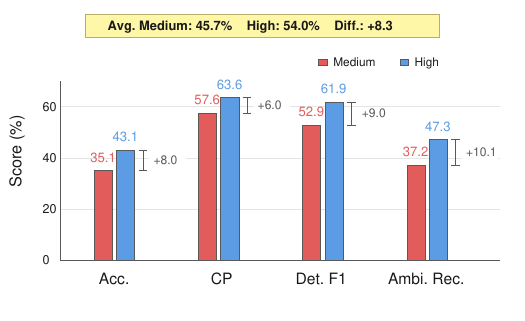}
    \caption{
    Reasoning-effort comparison for Doubao-Seed-2.0-Pro under neutral prompting.
    }
    \label{fig:reasoning_effort}
\end{figure}

Fig.~\ref{fig:reasoning_effort} shows that increasing reasoning effort leads to consistent overall improvements. Taking Doubao-Seed-2.0-Pro as an example, when the reasoning effort is increased from medium to high, the average score rises from 45.7\% to 54.0\%, with an overall gain of 8.3 points. The improvements are particularly pronounced on ambiguity-related metrics: Det. F1 increases by 9.0 points, and Ambi. Rec. improves from 37.2\% to 47.3\%, yielding a 10.1-point gain, which is larger than the improvement on CP. This suggests that higher reasoning effort mainly helps models identify ambiguous search states, compare multiple candidate entities, and incorporate user clues into subsequent search refinement.

This result is consistent with the characteristics of \textsc{DiscoBench}. Clarification-aware deep search requires models not only to retrieve the final answer, but also to continuously judge whether the evidence is sufficient across multiple ambiguity checkpoints, while maintaining and revising the search trajectory. However, even under the high-effort setting, the accuracy remains below 45\% and Ambi. Rec. remains below 50\%, indicating that simply increasing reasoning effort is still insufficient for robust clarification-aware behavior. Models still need stronger mechanisms for ambiguity localization, evidence verification, and deciding when to ask rather than directly guess.

\section{Additional Analysis by Ambiguity Complexity}
\label{app:complexity}

\begin{figure}[t]
    \centering
   \includegraphics[width=0.45\textwidth]{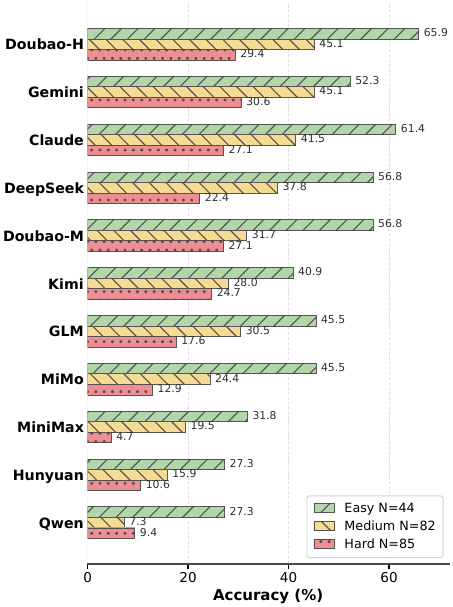}
    \caption{
    Performance across ambiguity-complexity levels under neutral prompting.
    }
    \label{fig:complexity_analysis}
\end{figure}

Fig.~\ref{fig:complexity_analysis} presents model performance across different levels of complexity of ambiguity under neutral prompting. Across nearly all evaluated models, accuracy consistently decreases from Easy to Hard, showing that ambiguity complexity introduces substantial additional difficulty beyond ordinary retrieval and reasoning.
Although stronger models such as Doubao-Seed-2.0-Pro (High), Gemini-3.1-Pro, Claude-Opus-4.7, and DeepSeek-V4-Pro achieve relatively strong performance on Easy examples, their accuracy still drops markedly on Hard examples. This suggests that increasing ambiguity complexity challenges not only evidence retrieval, but also the model’s ability to recognize underspecified states and proactively request clarification during multi-step search.

The performance gap between Easy and Hard settings further indicates that stronger reasoning ability alone is insufficient for robust clarification-aware search. As ambiguity becomes more subtle and accumulates across multiple checkpoints, models are increasingly prone to following plausible but incorrect search trajectories without initiating clarification. Lower-performing models exhibit the same downward trend from a lower baseline, indicating simultaneous weaknesses in both basic task completion and ambiguity resolution.

\section{Profile Classification Details}
\label{app:profile_details}

\paragraph{Behavioral profiles.}
We classify ambiguous-checkpoint trajectories into four profiles:
\begin{itemize}
    \item \textbf{DirectGuess}: no \textsc{Ask} with search count $\leq K$.
    \item \textbf{SearchHeavyGuess}: no \textsc{Ask} with search count $>K$.
    \item \textbf{DirectAsk}: ask before any retrieval.
    \item \textbf{SearchThenAsk}: retrieve before clarification.
\end{itemize}

The threshold $K\!=\!3$ is determined data-driven as the median search count among successful no-ask trajectories.

\paragraph{DirectAsk rarity.}
DirectAsk is extremely rare, accounting for only $0$--$7$ trajectories per model. For 7 of 9 models, the count satisfies $N\!\leq\!1$, indicating that current models almost never initiate clarification before retrieval.

\paragraph{Common Subset Robustness.}
\label{app:robustness}

The common subset contains 146 ambi-CPs reached by all 9 models in Tab.~\ref{tab:profile_pass}. GPT-5.4 is excluded due to prompt filtering, while Qwen is excluded because of insufficient reach. Model rankings on the common subset remain highly consistent with the full dataset (Spearman $\rho\!=\!0.95$), indicating that the observed behavioral patterns are robust to reach-rate differences.

\begin{table}[t]
    \centering
    \caption{
    \small
    Behavioral profile distribution (\%) on the common subset ($N\!=\!146$ ambi-CPs). Models are sorted by STA ratio.
    }
    \label{tab:profile_distribution}
    
    \footnotesize
    \setlength{\tabcolsep}{5pt}
    \renewcommand{\arraystretch}{1.05}
    
    \begin{tabular}{lcccc}
        \toprule
        \textbf{Model}
        & \textbf{DG}
        & \textbf{SHG}
        & \textbf{DA}
        & \textbf{STA} \\
        \midrule
        Gemini-3.1-Pro   & 24.7 & 15.8 & 4.8 & \textbf{54.8} \\
        Doubao-Seed-2.0-Pro  & 19.2 & 32.2 & 0.7 & \textbf{47.9} \\
        DeepSeek-V4-Pro & 26.0 & 41.8 & 0.7 & \textbf{31.5} \\
        Claude-Opus-4.7     & 41.1 & 27.4 & 0.7 & \textbf{30.8} \\
        MiMo-v2.5-Pro     & 19.2 & 51.4 & 0.0 & \textbf{29.5} \\
        GLM-5.1      & 41.1 & 30.8 & 0.0 & \textbf{28.1} \\
        MiniMax-M2.7   & 52.1 & 22.6 & 2.1 & \textbf{23.3} \\
        Kimi-K2.6     & 34.9 & 41.8 & 0.0 & \textbf{23.3} \\
        Hunyuan-3.0-Preview  & 24.0 & 52.7 & 0.7 & \textbf{22.6} \\
        \bottomrule
    \end{tabular}
    
    \vspace{1mm}
    
    {\scriptsize
    DG denotes DirectGuess; 
    SHG denotes SearchHeavyGuess; 
    DA denotes DirectAsk; 
    STA denotes SearchThenAsk.
    }
\end{table}

%
%

\section{Evaluated Models and API Configurations}
\label{app:model_details}

\begin{table}[!t]
\centering
\caption{%
Evaluated models and reasoning configurations.
\textbf{Reasoning} shorthand:
\emph{xhigh / high / medium} -- provider's discrete reasoning-effort or
thinking-level setting;
\emph{thinking} -- thinking/reasoning mode enabled
(no effort granularity exposed);
\emph{adapt.} -- adaptive thinking budget.%
}
\label{tab:model_details}
\scriptsize
\setlength{\tabcolsep}{3pt}
\renewcommand{\arraystretch}{1.15}
\begin{adjustbox}{max width=\columnwidth}
\begin{tabular}{@{}llll@{}}
\toprule
\textbf{Model (paper)} & \textbf{Provider} & \textbf{API Identifier} & \textbf{Reasoning} \\
\midrule
\rowcolor{clrRowA}
GPT-5.4                    & OpenAI      & \texttt{gpt-5.4-2026-03-05}         & xhigh           \\
\rowcolor{clrRowB}
Gemini-3.1-Pro-Preview     & Google      & \texttt{gemini-3.1-pro-preview}     & high            \\
\rowcolor{clrRowA}
Claude-Opus-4.7            & Anthropic   & \texttt{claude-opus-4-7}            & adapt.; max     \\
\midrule
\rowcolor{clrRowB}
Doubao-Seed-2.0-Pro-High   & ByteDance   & \texttt{doubao-seed-2-0-pro-260215} & high            \\
\rowcolor{clrRowA}
Doubao-Seed-2.0-Pro-Medium & ByteDance   & \texttt{doubao-seed-2-0-pro-260215} & medium          \\
\rowcolor{clrRowB}
DeepSeek-V4-Pro            & DeepSeek    & \texttt{deepseek-v4-pro}            & thinking; max   \\
\rowcolor{clrRowA}
Qwen3.6-Max       & Alibaba     & \texttt{qwen3.6-max-preview}        & thinking        \\
\rowcolor{clrRowB}
Kimi-K2.6         & Moonshot AI & \texttt{kimi-k2.6}                  & thinking        \\
\rowcolor{clrRowA}
MiMo-v2.5-Pro              & Xiaomi      & \texttt{mimo-v2.5-pro}              & thinking        \\
\rowcolor{clrRowB}
Hunyuan-3.0-Preview        & Tencent     & \texttt{hy3-preview}                & high            \\
\rowcolor{clrRowA}
MiniMax-M2.7               & MiniMax     & \texttt{MiniMax-M2.7}               & thinking        \\
\rowcolor{clrRowB}
GLM-5.1$^{\dagger}$        & Zhipu AI    & \texttt{z-ai/glm-5.1}               & thinking        \\
\midrule
\multicolumn{4}{@{}l}{\textit{User simulator and checkpoint judge}} \\
\midrule
\rowcolor{clrRowA}
Gemini-3-Flash-Medium      & Google      & \texttt{gemini-3-flash-preview}     & medium          \\
\bottomrule
\end{tabular}
\end{adjustbox}

\vspace{0.2em}
{\scriptsize $^{\dagger}$ GLM-5.1 is accessed via the OpenRouter gateway (\texttt{z-ai/glm-5.1}) rather than a direct Zhipu AI endpoint.}
\end{table}

Tab.~\ref{tab:model_details} lists the API configuration of every model
used in our main experiments and ablation studies, restricted to the
information needed to reproduce a call: the model name used in the
paper, the provider, the API identifier used at invocation time, and
the reasoning- or thinking-mode setting. To control for the confound of
reasoning budget, models exposing a configurable effort level are run
at the highest available setting in the main experiments;
Doubao-Seed-2.0-Pro-Medium is additionally included only for the
reasoning-effort analysis in Appendix~\ref{app:reasoning_effort}. Frontier proprietary models
(GPT-5.4, Gemini-3.1-Pro-Preview, and Claude-Opus-4.7) are restricted to
a concurrency of $5$ by provider-side rate limits, whereas the remaining
agent endpoints permit a concurrency of $300$; this asymmetry influences
wall-clock evaluation cost but not per-question correctness. All
\textsc{Search} calls are executed through Tavily regardless of the
agent backbone. A single auxiliary model, Gemini-3-Flash-Medium running
with thinking level set to \texttt{medium}, serves both as the simulated
user that releases discriminative clues during clarification turns and
as the checkpoint-level judge that scores each step; its configuration
is listed in the bottom block of Tab.~\ref{tab:model_details} and uses
a concurrency of $100$.

\section{Token Consumption}
\label{app:token_consumption}

To provide a supplementary reference for inference cost, we report the token consumption of the evaluated models in the main experiments and ablation studies. Tab.~\ref{tab:token_consumption} summarizes the total input and output tokens under the Neutral and Guided prompting settings. The Guided entry of GPT-5.4 is left blank because its Guided runs were excluded from the main analysis due to insufficient valid results. Doubao-Seed-2.0-Pro-Medium is included only for the reasoning-effort analysis under the Neutral setting, and therefore does not have a corresponding Guided entry.

Tab.~\ref{tab:ablation_token_consumption} further reports token consumption for the two ablation settings used in the main text: removing the search tool and evaluating on unambiguous questions. These ablation runs are conducted under neutral prompting and are reported separately from the main Neutral/Guided comparison.

\begin{table}[!t]
\centering
\caption{Token consumption of evaluated models under the Neutral and Guided prompting settings.}
\label{tab:token_consumption}
\scriptsize
\setlength{\tabcolsep}{3pt}
\renewcommand{\arraystretch}{1.15}
\begin{adjustbox}{max width=\columnwidth}
\begin{tabular}{@{}llrr@{}}
\toprule
\textbf{Model} & \textbf{Setting} & \textbf{Input} & \textbf{Output} \\
\midrule
Doubao-Seed-2.0-Pro-High
& Neutral & 1,250,346 & 11,891 \\
& Guided  & 2,343,428 & 17,068 \\

Gemini-3.1-Pro-Preview
& Neutral & 1,734,148 & 9,431 \\
& Guided  & 1,696,445 & 9,860 \\

Claude-Opus-4.7
& Neutral & 2,033,450 & 13,020 \\
& Guided  & 2,488,658 & 15,168 \\

DeepSeek-V4-Pro
& Neutral & 2,759,370 & 12,337 \\
& Guided  & 2,618,879 & 11,572 \\

Doubao-Seed-2.0-Pro-Medium
& Neutral & 957,009 & 7,038 \\
& Guided  & --      & -- \\

Kimi-K2.6
& Neutral & 975,818 & 6,073 \\
& Guided  & 1,956,450 & 10,574 \\

GLM-5.1
& Neutral & 556,046 & 4,869 \\
& Guided  & 1,165,076 & 8,343 \\

GPT-5.4
& Neutral & 3,554,723 & 22,536 \\
& Guided  & - & - \\

MiMo-v2.5-Pro
& Neutral & 1,458,475 & 7,298 \\
& Guided  & 2,148,970 & 9,847 \\

Hunyuan-3.0-Preview
& Neutral & 989,823 & 10,135 \\
& Guided  & 903,426 & 9,295 \\

MiniMax-M2.7
& Neutral & 311,847 & 5,413 \\
& Guided  & 367,398 & 5,631 \\

Qwen3.6-Max
& Neutral & 1,409,768 & 9,986 \\
& Guided  & 3,697,178 & 17,444 \\
\bottomrule
\end{tabular}
\end{adjustbox}

\end{table}

\begin{table}[!t]
\centering
\caption{Token consumption in the ablation studies under neutral prompting.}
\label{tab:ablation_token_consumption}
\scriptsize
\setlength{\tabcolsep}{3pt}
\renewcommand{\arraystretch}{1.15}
\begin{adjustbox}{max width=\columnwidth}
\begin{tabular}{@{}lrrrr@{}}
\toprule
\multirow{2}{*}{\textbf{Model}} 
& \multicolumn{2}{c}{\textbf{w/o Search}} 
& \multicolumn{2}{c}{\textbf{Unambig. Qs.}} \\
\cmidrule(lr){2-3}\cmidrule(lr){4-5}
& \textbf{Input} & \textbf{Output} 
& \textbf{Input} & \textbf{Output} \\
\midrule
Doubao-Seed-2.0-Pro-High & 54,867 & 11,619 & 1,640,655 & 13,125 \\
Gemini-3.1-Pro-Preview   & 24,030 & 10,810 & 1,575,854 & 9,761  \\
DeepSeek-V4-Pro          & 8,109  & 14,281 & 4,328,007 & 8,520  \\
Hunyuan-3.0-Preview      & 65,534 & 19,870 & 1,505,374 & 11,648 \\
MiniMax-M2.7             & 3,105  & 2,820  & 1,500,414 & 18,199 \\
\bottomrule
\end{tabular}
\end{adjustbox}
\end{table}

\section{Reproducibility under a Black-Box Search Backend}
\label{sec:appendix-reproducibility}

All \textsc{Search} calls in \textsc{DiscoBench} are routed through
Tavily~\cite{tavily2026}, a hosted web-search API whose index,
ranking model, and freshness policies are not publicly disclosed. From
the agent's point of view the backend therefore behaves as a black
box: the same query issued on different days can return different
snippet sets, different rankings, and even different source domains,
both because the web itself is non-stationary (pages appear, are
edited, or are de-indexed) and because Tavily's own retrieval and
re-ranking stack can be updated without notice. Strict bit-exact
reproducibility of an end-to-end trajectory is therefore not
achievable.

\paragraph{Solvability of individual instances.}
This stochasticity does not, however, undermine the solvability of
individual questions, because the gold answer in \textsc{DiscoBench}
is by construction time-invariant. Every question is built on stable,
verifiable factual knowledge, so the correct answer does not drift as
the web changes. Tavily's black-box behavior varies only the
\emph{retrieval surface}: which snippet from which source is returned,
and in what order. Since the underlying evidence lives in
well-established, widely indexed public sources, any reasonably
comprehensive web index can be expected to surface at least one
supporting snippet for a reasonable query. \textsc{DiscoBench}
therefore remains solvable in principle on every run.

\paragraph{Variance across runs.}
The black-box effect manifests not as a change in answerability but as
variance in \emph{trajectory shape and retrieval efficiency}: the
specific snippets surfaced, their ranking, and therefore which queries
are sufficient and how many \textsc{Search} calls an agent needs
before the relevant evidence appears in its context window. Two runs
of the same agent on the same question may consequently take different
paths and consume different numbers of tool calls even when both
ultimately succeed.

\paragraph{Implications for interpretation and replication.}
A model's reported score on \textsc{DiscoBench} should be read as an
expectation over Tavily snapshots rather than as a per-run
deterministic quantity, and an individual incorrect trajectory should
be inspected for its underlying cause---retrieval ordering on a given
day versus a genuine ambiguity-handling failure---before being
attributed to model capability. To make replications maximally
comparable, we recommend that future users of \textsc{DiscoBench}
(i)~run all compared agents within a short, contiguous evaluation
window so they observe near-identical Tavily snapshots, and
(ii)~where feasible, cache and release the raw Tavily responses
alongside model outputs, which converts a hard reproducibility problem
(re-deriving an identical live web view) into a tractable one
(replaying a fixed snippet log).

\section{Examples}
\label{app:examples}
 
To illustrate the design of \textsc{\textsc{DiscoBench}}, we walk through three
cases from the benchmark, each spotlighting a different ambiguity type.
Case~1 (Tab.~\ref{tab:case-entity-cascade}) shows cascading Entity
ambiguity along a multi-hop chain; Case~2
(Tab.~\ref{tab:case-factinacc-entity}) pairs Factual Inaccuracy
(CP$_1$) with Version ambiguity (CP$_3$) in a single trajectory,
forcing the agent to switch detection strategies mid-chain; Case~3
(Tab.~\ref{tab:case-criteria}) presents Criteria ambiguity, where
identical surface wording maps to two distinct ranking standards. Each
case lists the original question, the ambiguity-injected rewrite shown
to the agent, and the checkpoint trajectory with ground-truth types,
targets, ambiguity logic, and the discriminative clue the user
simulator releases upon a well-targeted \textsc{Ask}.
In every table, English rows (gray tint) precede the original Chinese
(cream tint). Colored text tracks each substitution thread: the same
color appears on the original constraint, its weakened rewrite, and
the clue that later restores it.

\paragraph{Case 1: cascading Entity ambiguity across a multi-hop chain.}
Tab.~\ref{tab:case-entity-cascade} presents a four-checkpoint question
from the \emph{Video Games} domain. The original question fully
specifies two distinguishing awards -- the
\subA{GWB Indie Game Awards Bronze Prize} and the
\subB{2012 Top Ten Most Anticipated Web Games} -- which uniquely pin
down the two target entities \emph{My Time at Sandrock} (a Pathea
Games title) and \emph{Jingtian Zhanshen Online} (a ZQ~Game title).
The ambiguity-injection step replaces each of these strong constraints
with the generic phrase ``\emph{won an award}'', producing two
cascading Ambi checkpoints (CP$_1$ and CP$_3$) of type \emph{Entity}:
at each, the rewritten constraint matches multiple notable
award-winning products of the queried company, so an unaided
\textsc{Search} returns more than one candidate. The agent should
detect this and invoke \textsc{Ask}; the user
simulator then releases the original award name as a discriminative
clue, allowing a refined \textsc{Search} to lock onto the target.
This case also exposes the benchmark's central failure mode,
\emph{silent cascading}: an incorrect resolution at
CP$_1$ (e.g., \emph{Portia} instead of \emph{Sandrock}) still routes
to a syntactically valid publisher at CP$_2$, but every downstream
checkpoint then targets the wrong entity, with no local indication of
the upstream error.

\paragraph{Case 2: Factual Inaccuracy followed by Version ambiguity.}
Tab.~\ref{tab:case-factinacc-entity} presents a four-checkpoint
question from the \emph{Sports} domain. Two substitution threads
run through the trajectory but follow different injection patterns. Thread~A (CP$_1$) is a \emph{Factual Inaccuracy}: the
original country nickname ``\subA{Land of Windmills}'' (the
Netherlands) is replaced by ``\subA{Land of Hajimi}'', a fabricated
term that does not refer to any real country. The agent cannot resolve
this checkpoint through retrieval alone -- a faithful search returns
no match -- and must invoke \textsc{Ask} rather than guess. Thread~B
(CP$_3$) is a \emph{Version} ambiguity: the original match identifier
``\subB{at the 60th minute of a 2018 CSL match}'' is weakened to the
looser window ``\subB{in a CSL match in March--April}''. Within that
window Wang~Chu came on as a substitute in two different matches on
different dates, replacing Wang~Gang in one and Cao~Yongjing in the
other; only the precise date disambiguates which match is meant. The
two threads together stress that \textsc{DiscoBench} requires the agent to
switch \emph{detection mode} within a single trajectory rather than
apply a single clarification heuristic uniformly.

\paragraph{Case 3: Criteria ambiguity in a long bridging chain.}
Tab.~\ref{tab:case-criteria} presents a three-checkpoint question
spanning the \emph{Technology} domain. Unlike Case~1 and
Case~2, here a single colored phrase in the original is \emph{deleted}
rather than \emph{replaced}: the qualifier
``\subA{also nicknamed the `Ice City'}'' is removed during
ambiguity injection.
After deletion, the surviving constraint -- ``a city whose Chinese
name has three characters and lies above 40$^{\circ}$N, listed among
the top three [beer-festival] cities'' -- can be satisfied under two
distinct enumeration criteria: the world's top three beer festivals
(yielding Munich, \zh{慕尼黑}, $\sim$48$^{\circ}$N) or China's top
three beer festivals (yielding Harbin, \zh{哈尔滨},
$\sim$45$^{\circ}$N). The agent must recognize that a single
description fits two rankings and clarify \emph{which ranking} the
user intends before proceeding.
 
\begin{table}[!tbp]
\centering
\caption{%
\textbf{Case~1 (Entity, cascading).} Four-checkpoint question
from the \emph{Video Games} domain; two cascading Entity-type Ambi
checkpoints (CP$_1$, CP$_3$). Text color marks the two substitution
threads.%
}
\label{tab:case-entity-cascade}
\scriptsize
\setlength{\tabcolsep}{3pt}
\renewcommand{\arraystretch}{1.05}
\begin{tabular}{@{}
  >{\raggedright\arraybackslash}p{0.10\columnwidth}
  >{\raggedright\arraybackslash}p{0.85\columnwidth}
@{}}
\toprule
 
\multicolumn{2}{@{}p{0.96\columnwidth}@{}}{%
  \textbf{Domain:} Video Games \;\textbar\;
  \textbf{Difficulty:} Medium \;\textbar\;
  \textbf{\#~CPs:} 4 \;\textbar\;
  \textbf{\#~Ambi:} 2 (Entity) \;\textbar\;
  \textbf{Gold:} \texttt{2013/6/26}%
} \\
\midrule
 
\rowcolor{groupBg}
\multicolumn{2}{@{}p{0.96\columnwidth}@{}}{%
  \textsc{\textbf{Part~1\,---\,Input question pair}}%
  \;\;(original seed and its ambiguity-injected rewrite)%
} \\
\midrule
 
\multicolumn{2}{@{}p{0.96\columnwidth}@{}}{%
  \textbf{Original Q} \,\textit{(seed, before ambiguity injection; reference only)}%
} \\
\rowcolor{enBg}
\textsf{en} & As of February~2026, a representative work of Chongqing
              Pathea Games won the \subA{GWB Indie Game Awards Bronze
              Prize}; this work is the sequel to a certain game whose
              publisher has another game that won the
              \subB{2012 Top Ten Most Anticipated Web Games}; when was
              the first batch of licensed agents of that latter game
              (in June~2012) formally confirmed to be acquired? \\
\rowcolor{zhBg}
\textsf{zh} & \zh{截止2026年2月份，重庆帕斯亚科技有限公司的某部代表作品获\subA{GWB独立游戏大奖赛铜奖}，这个代表作品是某游戏的续作，这个游戏的出版单位有一款游戏荣获\subB{2012年度最受玩家期待的十大网页游戏}，这款游戏2012年6月的首批代理什么时间正式确定被收购？} \\
\midrule
 
\multicolumn{2}{@{}p{0.96\columnwidth}@{}}{%
  \textbf{Rewritten Q} \,\textit{(after ambiguity injection; shown to agent)}%
} \\
\rowcolor{enBg}
\textsf{en} & As of February~2026, a representative work of Chongqing
              Pathea Games \subA{won an award}; this work is the sequel
              to a certain game whose publisher has another game that
              \subB{won an award}; when was the first batch of licensed
              agents of that latter game (in June~2012) formally
              confirmed to be acquired? \\
\rowcolor{zhBg}
\textsf{zh} & \zh{截止2026年2月份，重庆帕斯亚科技有限公司的某部代表作品\subA{获奖}，这个代表作品是某游戏的续作，这个游戏的出版单位有一款游戏\subB{获奖}，这款游戏2012年6月的首批代理什么时间正式确定被收购？} \\
\midrule
 
\rowcolor{groupBg}
\multicolumn{2}{@{}p{0.96\columnwidth}@{}}{%
  \textsc{\textbf{Part~2\,---\,Checkpoint trajectory}}%
  \;\;(four sub-questions, evaluated in order)%
} \\
\midrule
 
\multicolumn{2}{@{}p{0.96\columnwidth}@{}}{%
  \textbf{CP$_1$} \,---\, \textbf{Ambi} \textit{(Entity)}%
} \\
\rowcolor{enBg}
\textbf{Sub-Q} & Which representative work of Pathea Games \subA{won an award}? \\
\rowcolor{zhBg}
               & \zh{重庆帕斯亚科技有限公司有哪部代表作品\subA{获奖}？} \\
\rowcolor{enBg}
\textbf{Target} & My Time at Sandrock \\
\rowcolor{zhBg}
                & \zh{《沙石镇时光》} \\
\rowcolor{enBg}
\textbf{Logic} & Two Pathea Games works satisfy ``\subA{won an award}'':
                 My Time at Portia, My Time at Sandrock. \\
\rowcolor{zhBg}
               & \zh{帕斯亚科技有两部代表获奖作品 ——《波西亚时光》、《沙石镇时光》——均满足``\subA{获奖}''约束。} \\
\rowcolor{enBg}
\textbf{Clue}  & ``I remember it won the \subA{GWB Indie Game Awards Bronze Prize}.'' \\
\rowcolor{zhBg}
               & \zh{``我记得是获\subA{GWB独立游戏大奖赛铜奖}''} \\
\midrule
 
\multicolumn{2}{@{}p{0.96\columnwidth}@{}}{%
  \textbf{CP$_2$} \,---\, Unambi%
} \\
\rowcolor{enBg}
\textbf{Sub-Q} & This work is the sequel to which game, and who is its publisher? \\
\rowcolor{zhBg}
               & \zh{这个代表作品是某游戏的续作，这个游戏的出版单位是？} \\
\rowcolor{enBg}
\textbf{Target} & Shenzhen ZQ~Game Co., Ltd. \\
\rowcolor{zhBg}
                & \zh{深圳中青宝互动网络股份有限公司} \\
\midrule
 
\multicolumn{2}{@{}p{0.96\columnwidth}@{}}{%
  \textbf{CP$_3$} \,---\, \textbf{Ambi} \textit{(Entity)}%
} \\
\rowcolor{enBg}
\textbf{Sub-Q} & Among ZQ~Game's products, which game has \subB{won an award}? \\
\rowcolor{zhBg}
               & \zh{深圳中青宝互动网络股份有限公司的产品中哪个游戏曾\subB{获奖}？} \\
\rowcolor{enBg}
\textbf{Target} & Jingtian Zhanshen Online \\
\rowcolor{zhBg}
                & \zh{《惊天战神 OL》} \\
\rowcolor{enBg}
\textbf{Logic} & Three ZQ~Game titles satisfy ``\subB{won an award}'':
                 War of Resistance, Drawn Sword, Jingtian Zhanshen Online. \\
\rowcolor{zhBg}
               & \zh{中青宝有三款代表获奖游戏 ——《抗战》、《亮剑》、《惊天战神 OL》—— 均满足``\subB{获奖}''约束。} \\
\rowcolor{enBg}
\textbf{Clue}  & ``I remember it won the \subB{2012 Top Ten Most Anticipated Web Games award}.'' \\
\rowcolor{zhBg}
               & \zh{``我记得是荣获\subB{2012年度最受玩家期待的十大网页游戏奖}''} \\
\midrule
 
\multicolumn{2}{@{}p{0.96\columnwidth}@{}}{%
  \textbf{CP$_4$} \,---\, Unambi \textit{(final)}%
} \\
\rowcolor{enBg}
\textbf{Sub-Q} & When was the first batch of licensed agents of this game (in June~2012) formally confirmed to be acquired? \\
\rowcolor{zhBg}
               & \zh{这款游戏2012年6月的首批代理什么时间正式确定被收购？} \\
\multicolumn{2}{@{}p{0.96\columnwidth}@{}}{%
  \textbf{Target:} \texttt{2013/6/26}%
} \\
 
\bottomrule
\end{tabular}
\end{table}

\begin{table}[!tbp]
\centering
\caption{%
\textbf{Case~2 (Factual Inaccuracy + Version).} Four-checkpoint
question from the \emph{Sports} domain; CP$_1$ injects a fabricated
country nickname, CP$_3$ injects an under-specified timing window
admitting two candidate teammates.%
}
\label{tab:case-factinacc-entity}
\scriptsize
\setlength{\tabcolsep}{3pt}
\renewcommand{\arraystretch}{1.05}
\begin{tabular}{@{}
  >{\raggedright\arraybackslash}p{0.10\columnwidth}
  >{\raggedright\arraybackslash}p{0.85\columnwidth}
@{}}
\toprule
 
\multicolumn{2}{@{}p{0.96\columnwidth}@{}}{%
  \textbf{Domain:} Sports \;\textbar\;
  \textbf{Difficulty:} Medium \;\textbar\;
  \textbf{\#~CPs:} 4 \;\textbar\;
  \textbf{\#~Ambi:} 2 (Factual~Inaccuracy, Version) \;\textbar\;
  \textbf{Gold:} \texttt{Chongqing}%
} \\
\midrule
 
\rowcolor{groupBg}
\multicolumn{2}{@{}p{0.96\columnwidth}@{}}{%
  \textsc{\textbf{Part~1\,---\,Input question pair}}%
  \;\;(original seed and its ambiguity-injected rewrite)%
} \\
\midrule
 
\multicolumn{2}{@{}p{0.96\columnwidth}@{}}{%
  \textbf{Original Q} \,\textit{(seed, before ambiguity injection; reference only)}%
} \\
\rowcolor{enBg}
\textsf{en} & A certain country is known as the
              \subA{``Land of Windmills''}. A city football team is
              based in this country's seat of government. In 2013, a
              Chinese player trial-trained with that team; in a 2018
              Chinese Super League (CSL) match, this player came on as
              a substitute for a male athlete
              \subB{at the 60th minute}. Where was that male athlete
              born? \\
\rowcolor{zhBg}
\textsf{zh} & \zh{某个国家被称为``\subA{风车之国}''，该国家的政府所在地有一个城市球队。2013年一名中国球员在该队试训，该球员在\subB{2018年中超比赛中第60分钟}替补一名男运动员出场，该男运动员出生于哪里？} \\
\midrule
 
\multicolumn{2}{@{}p{0.96\columnwidth}@{}}{%
  \textbf{Rewritten Q} \,\textit{(after ambiguity injection; shown to agent)}%
} \\
\rowcolor{enBg}
\textsf{en} & A certain country is known as the
              \subA{``Land of Hajimi''}. A city football team is based
              in this country's seat of government. In 2013, a Chinese
              player trial-trained with that team; in
              \subB{a CSL match in March--April}, this player came on
              as a substitute for a male athlete. Where was that male
              athlete born? \\
\rowcolor{zhBg}
\textsf{zh} & \zh{某个国家被称为``\subA{哈吉米之国}''，该国家的政府所在地有一个城市球队。2013年有中国球员在该队试训，该球员在\subB{3月～4月中超的某个比赛中}替补一名男运动员出场，该男运动员出生于哪里？} \\
\midrule
 
\rowcolor{groupBg}
\multicolumn{2}{@{}p{0.96\columnwidth}@{}}{%
  \textsc{\textbf{Part~2\,---\,Checkpoint trajectory}}%
  \;\;(four sub-questions, evaluated in order)%
} \\
\midrule
 
\multicolumn{2}{@{}p{0.96\columnwidth}@{}}{%
  \textbf{CP$_1$} \,---\, \textbf{Ambi} \textit{(Factual~Inaccuracy)}%
} \\
\rowcolor{enBg}
\textbf{Sub-Q} & Which country is known as the \subA{``Land of Hajimi''}? \\
\rowcolor{zhBg}
               & \zh{哪个国家被称为``\subA{哈吉米之国}''？} \\
\rowcolor{enBg}
\textbf{Target} & the Netherlands \\
\rowcolor{zhBg}
                & \zh{荷兰} \\
\rowcolor{enBg}
\textbf{Logic} & \subA{``Land of Hajimi''} is not an actual nickname
                 for any country; it is a fabricated reference
                 deliberately injected as a factual inaccuracy.
                 Searches return no plausible match, so the agent must
                 invoke \textsc{Ask} for clarification rather than
                 commit to a guess. \\
\rowcolor{zhBg}
               & \zh{``\subA{哈吉米之国}''并非任何国家的真实别称，是出题阶段刻意杜撰的事实错误。retrieval 不会返回匹配结果，agent 应识别此事实错误并向用户求助，而非凭借猜测作答。} \\
\rowcolor{enBg}
\textbf{Clue}  & ``I misremembered -- it's the \subA{Land of Windmills}.'' \\
\rowcolor{zhBg}
               & \zh{``我记错了，是\subA{风车之国}''} \\
\midrule
 
\multicolumn{2}{@{}p{0.96\columnwidth}@{}}{%
  \textbf{CP$_2$} \,---\, Unambi%
} \\
\rowcolor{enBg}
\textbf{Sub-Q} & A city football team is based in this country's seat
                 of government; in 2013, a Chinese player trial-trained
                 with that team. Who is the player? \\
\rowcolor{zhBg}
               & \zh{该国家的政府所在地有一个城市球队，2013年有位中国球员在该队试训，该球员是？} \\
\rowcolor{enBg}
\textbf{Target} & Wang Chu \\
\rowcolor{zhBg}
                & \zh{王楚} \\
\midrule
 
\multicolumn{2}{@{}p{0.96\columnwidth}@{}}{%
  \textbf{CP$_3$} \,---\, \textbf{Ambi} \textit{(Version)}%
} \\
\rowcolor{enBg}
\textbf{Sub-Q} & In \subB{a CSL match in March--April 2018}, this
                 player came on as a substitute for a male athlete.
                 Who is that athlete? \\
\rowcolor{zhBg}
               & \zh{该球员\subB{2018年3月～4月中超的某个比赛中}替补一名男运动员出场，这个男运动员是？} \\
\rowcolor{enBg}
\textbf{Target} & Cao Yongjing \\
\rowcolor{zhBg}
                & \zh{曹永竞} \\
\rowcolor{enBg}
\textbf{Logic} & Within \subB{March--April 2018}, Wang Chu came on as
                 a substitute in two distinct CSL matches on different
                 dates, replacing Wang Gang and Cao Yongjing
                 respectively; the rewritten window matches both, and
                 only the exact date pins down which. \\
\rowcolor{zhBg}
               & \zh{王楚在\subB{2018年3月～4月}期间替补出场两次，在两个不同日期的中超比赛中分别替换了王刚和曹永竞；重写后的时间窗口同时匹配这两场，唯有具体日期才能锁定其中一场。} \\
\rowcolor{enBg}
\textbf{Clue}  & ``\subB{It was on April 8, 2018.}'' \\
\rowcolor{zhBg}
               & \zh{``\subB{是在2018年4月8日上场的}''} \\
\midrule
 
\multicolumn{2}{@{}p{0.96\columnwidth}@{}}{%
  \textbf{CP$_4$} \,---\, Unambi \textit{(final)}%
} \\
\rowcolor{enBg}
\textbf{Sub-Q} & Where was this athlete born? \\
\rowcolor{zhBg}
               & \zh{这个运动员出生于哪里？} \\
\multicolumn{2}{@{}p{0.96\columnwidth}@{}}{%
  \textbf{Target:} \texttt{Chongqing} \,/\, \zh{重庆市}%
} \\
 
\bottomrule
\end{tabular}
\end{table}

\begin{table}[!tbp]
\centering
\caption{%
\textbf{Case~3 (Criteria).} Three-checkpoint question in the
\emph{Technology} domain; one Criteria-type Ambi checkpoint
(CP$_2$). The substitution thread here is \emph{removed} rather than
replaced, so the colored phrase appears only in the original question
and in the discriminative clue.%
}
\label{tab:case-criteria}
\scriptsize
\setlength{\tabcolsep}{3pt}
\renewcommand{\arraystretch}{1.05}
\begin{tabular}{@{}
  >{\raggedright\arraybackslash}p{0.10\columnwidth}
  >{\raggedright\arraybackslash}p{0.85\columnwidth}
@{}}
\toprule
 
\multicolumn{2}{@{}p{0.96\columnwidth}@{}}{%
  \textbf{Domain:} Technology \;\textbar\;
  \textbf{Difficulty:} Easy \;\textbar\;
  \textbf{\#~CPs:} 3 \;\textbar\;
  \textbf{\#~Ambi:} 1 (Criteria) \;\textbar\;
  \textbf{Gold:} \texttt{12}%
} \\
\midrule
 
\rowcolor{groupBg}
\multicolumn{2}{@{}p{0.96\columnwidth}@{}}{%
  \textsc{\textbf{Part~1\,---\,Input question pair}}%
  \;\;(original seed and its ambiguity-injected rewrite)%
} \\
\midrule
 
\multicolumn{2}{@{}p{0.96\columnwidth}@{}}{%
  \textbf{Original Q} \,\textit{(seed, before ambiguity injection; reference only)}%
} \\
\rowcolor{enBg}
\textsf{en} & The R\&D of a certain manned submersible was designated
              as a key special project under the National 863~Program
              in 2002. The submersible made its first dives in the
              world's third-largest ocean in a certain year; in April
              of that same year, a research vessel completed its
              maiden voyage from a certain dock. That dock is a
              sub-project of a certain plaza, which opens to visitors
              free of charge during a certain folk festival. The
              festival is associated with a ``top three [cities]''
              ranking; among the three cities, one has a Chinese name
              consisting of three characters and lies above
              40$^{\circ}$N \subA{and is also nicknamed the
              ``Ice City''}. A satellite launched from this city on
              9~June~2023 was developed by a certain university. As of
              April~2025, how many sci-tech innovation teams does that
              university have? \\
\rowcolor{zhBg}
\textsf{zh} & \zh{某载人潜水器的研制工作在2002年被列为863计划重大专项，该载人潜水器在某年首次奔赴世界第三大洋开展下潜作业，同年4月，某考察船在某码头完成首次启航，该码头是某广场的子项目，该广场在某一节日期间免费向游人开放，该民间节日有一个榜单，榜单的三大城市中有一个三字城市位于北纬40多度\subA{且别称是``冰城''}，该城市2023年6月9日发射的卫星由一所学校研制，该学校截止2025年4月，科技创新团队有多少个？} \\
\midrule
 
\multicolumn{2}{@{}p{0.96\columnwidth}@{}}{%
  \textbf{Rewritten Q} \,\textit{(after ambiguity injection; shown to agent)}%
} \\
\rowcolor{enBg}
\textsf{en} & Identical to the Original~Q above, with the colored
              fragment \subA{``and is also nicknamed the `Ice City'\,''}
              \emph{deleted}. \\
\rowcolor{zhBg}
\textsf{zh} & \zh{与上方原始问句完全一致，仅删除标红片段 \subA{``且别称是`冰城'\,''}。} \\
\midrule
 
\rowcolor{groupBg}
\multicolumn{2}{@{}p{0.96\columnwidth}@{}}{%
  \textsc{\textbf{Part~2\,---\,Checkpoint trajectory}}%
  \;\;(three sub-questions, evaluated in order)%
} \\
\midrule
 
\multicolumn{2}{@{}p{0.96\columnwidth}@{}}{%
  \textbf{CP$_1$} \,---\, Unambi%
} \\
\rowcolor{enBg}
\textbf{Sub-Q} & (Chained from the early constraints:) the dock is a
                 sub-project of a plaza that opens free of charge
                 during which folk festival? \\
\rowcolor{zhBg}
               & \zh{某载人潜水器的研制工作在2002年被列为863计划重大专项，该载人潜水器在某年首次奔赴世界第三大洋开展下潜作业，同年4月，某考察船在某码头完成首次启航，该码头是某广场的子项目，该广场在某一节日期间免费向游人开放，这个节日是？} \\
\rowcolor{enBg}
\textbf{Target} & the Beer Festival (\zh{啤酒节}) \\
\rowcolor{zhBg}
                & \zh{啤酒节} \\
\midrule
 
\multicolumn{2}{@{}p{0.96\columnwidth}@{}}{%
  \textbf{CP$_2$} \,---\, \textbf{Ambi} \textit{(Criteria)}%
} \\
\rowcolor{enBg}
\textbf{Sub-Q} & The folk festival is associated with a ``top three
                 [cities]'' ranking; one of the three cities has a
                 Chinese name of three characters and lies above
                 40$^{\circ}$N. Which ranking is being referred to? \\
\rowcolor{zhBg}
               & \zh{该民间节日有一个榜单，榜单中有一个三字城市位于北纬40多度，这个榜单是什么？} \\
\rowcolor{enBg}
\textbf{Target} & China's Top Three Beer Festivals \\
\rowcolor{zhBg}
                & \zh{中国三大啤酒节} \\
\rowcolor{enBg}
\textbf{Logic} & ``Top Three Beer Festivals'' admits two distinct
                 enumeration criteria: \emph{the world's} top three
                 versus \emph{China's} top three. Each list contains a
                 city whose Chinese name has three characters and lies
                 above 40$^{\circ}$N --- Munich
                 (\zh{慕尼黑}, $\sim$48$^{\circ}$N) in the world list,
                 Harbin (\zh{哈尔滨}, $\sim$45$^{\circ}$N) in the
                 China list. The rewritten constraint alone cannot
                 tell the two lists apart. \\
\rowcolor{zhBg}
               & \zh{``三大啤酒节''这一榜单的口径存在二义性：可指``世界三大啤酒节''或``中国三大啤酒节''。两份榜单各自包含一座中文名为三字、纬度在北纬40°以上的城市——世界榜中的慕尼黑（约48°N）与中国榜中的哈尔滨（约45°N）。重写后的约束本身无法区分这两份榜单。} \\
\rowcolor{enBg}
\textbf{Clue}  & ``The three-character city is also nicknamed the
                 \subA{`Ice City'}.'' \\
\rowcolor{zhBg}
               & \zh{``三字城市别称是\subA{`冰城'}''} \\
\midrule
 
\multicolumn{2}{@{}p{0.96\columnwidth}@{}}{%
  \textbf{CP$_3$} \,---\, Unambi \textit{(final)}%
} \\
\rowcolor{enBg}
\textbf{Sub-Q} & A satellite launched from Harbin on 9~June~2023 was
                 developed by a certain university. As of April~2025,
                 how many sci-tech innovation teams does that
                 university have? \\
\rowcolor{zhBg}
               & \zh{哈尔滨2023年6月9日发射的卫星由一所学校研制，该学校截止2025年4月，科技创新团队有多少个？} \\
\multicolumn{2}{@{}p{0.96\columnwidth}@{}}{%
  \textbf{Target:} \texttt{12}%
} \\
 
\bottomrule
\end{tabular}
\end{table}
 
\FloatBarrier

\section{Annotation Details}
\label{sec:appendix-annotation}
\paragraph{Recruitment and compensation.}
Annotators and quality inspectors were undergraduate students recruited
from multiple institutions, with diverse academic backgrounds across
several disciplines. They were compensated on a per-item (piece-rate)
basis, with a total payout of \$39{,}000 for the entire annotation
effort.

\paragraph{Annotator consent.}
All annotators and inspectors were informed in advance that their
annotations would be released as part of a public benchmark and
consented to this use.

\paragraph{Ethics review.}
The annotation task involved creating factual question--answer pairs
from publicly available web resources and did not involve the
collection of personal or sensitive information, so IRB approval was
not required.

\section{Quality Inspection}
\label{app:quality-inspection}

This section provides further details on the quality control (QC) process introduced in Section~\ref{sec:data_quality}.
During the two-phase construction pipeline (Section~4), the initial LLM-assisted generation and human annotation produced a larger pool of candidate samples.
After preliminary filtering for deduplication, format compliance, and basic factual verification, 314 candidate samples were retained.
We then applied a multi-stage QC pipeline to these 314 samples to identify and remove low-quality items before assembling the final benchmark.
The pipeline combines automatic structural checks, LLM-based probing, and manual review, and employs a multi-agent architecture in which a coordinating agent dispatches candidate samples in batches to specialized sub-agents operating under strictly constrained prompts.

\paragraph{Stage~1: Structural Validation.}
An automatic script verifies every candidate sample for field completeness (all required fields non-empty), checkpoint-structure consistency (each sample contains at least one \textit{Ambi} and one terminal checkpoint; every \textit{Ambi} node carries a non-empty \texttt{ambiguity\_logic} and \texttt{clue\_if\_asked}), and difficulty--label alignment (Easy / Medium / Hard corresponds to 1 / 2 / 3 ambiguity checkpoints, respectively).

\paragraph{Stage~2: LLM-Based Probing.}
Each sample is independently tested under two complementary conditions to assess whether the task genuinely requires multi-step retrieval and multi-turn clarification.
In both cases, the sub-agent receives the \emph{complete rewritten question} (i.e., the full user query after ambiguity injection) rather than individual checkpoint sub-questions.
The prompt templates are provided in Box~\ref{box:qc-prompts}.

\begin{itemize}
\item \textbf{Closed-book probing.}
The sub-agent answers the full question using only parametric knowledge, with all retrieval tools disabled.
A correct answer signals potential \textit{knowledge leakage}.
Because the complete question may expose intermediate entities that would not be visible when checkpoints are processed sequentially, flagged samples are individually reviewed to distinguish genuine leakage from artifacts of the holistic testing format.

\item \textbf{Open-book probing.}
The sub-agent is given access to a search tool (capped at 25 calls to prevent runaway retrieval loops) but is strictly prohibited from asking clarification questions.
A correct answer under this constraint signals \textit{clarification-free solvability}: the injected ambiguity may not effectively require multi-turn clarification.
\end{itemize}

Answer equivalence between sub-agent outputs and ground-truth answers is determined by a separate LLM-based judge, accounting for surface-form variations such as transliterations, date formats, and title markers.

\paragraph{Stage~3: Ambiguity and Factual-Error Assessment.}
For each ambiguous checkpoint, a sub-agent assesses whether the ambiguity is \textit{surface-level}, i.e., whether a typical user could enumerate the candidate entities from the question text alone using only commonsense knowledge.
Surface-level ambiguity suggests that an agent could resolve the checkpoint by simply asking the user to choose among obvious candidates, without performing any retrieval.
Separately, for checkpoints of the \textit{Factual Inaccuracy} type, the sub-agent evaluates whether the injected error is recognizable without retrieval.
Errors that a typical user could identify through commonsense alone (e.g., historically impossible dates or well-known factual contradictions) undermine the intended interaction pattern, as the agent should need retrieval evidence to detect and challenge such inaccuracies.

\paragraph{Stage~4: Manual Review and Answer Verification.}
All automatically flagged samples undergo manual review covering three aspects:
(1)~\textit{question and ambiguity design}, including whether the question text uniquely constrains the expected answer, whether the injected ambiguity is realistically triggerable during retrieval, and whether sub-questions are logically coherent with the overall reasoning chain;
(2)~\textit{clue and retrieval quality}, including whether the discriminative clue is natural and sufficient for disambiguation, and whether the target answer is retrievable through mainstream search engines;
and (3)~\textit{answer correctness}, where we cross-check ground-truth annotations against external sources.
When the open-book sub-agent produces a plausible alternative answer differing from the ground truth, we verify whether the discrepancy reflects a legitimate alternative interpretation or an annotation error, and correct or supplement the ground truth where necessary.
Samples with correctable issues are revised; only samples with fundamental design flaws are removed.

\paragraph{Stage~5: Final Verdict.}
A rule-based decision tree aggregates the signals from the preceding stages to determine whether each sample effectively requires both deep retrieval and multi-turn clarification.
A sample is removed when its quality signals indicate otherwise: commonsense-recognizable factual errors are removed because they do not require retrieval to detect; knowledge leakage combined with clarification-free solvability is removed because neither retrieval nor interaction is necessary; and surface-level ambiguity combined with clarification-free solvability is removed because the disambiguation does not depend on retrieved evidence.
Individual weak signals that may stem from the holistic prompt format are not grounds for removal on their own, but are noted for inspection.

\paragraph{Results.}
Tab.~\ref{tab:qc-results} summarizes the QC outcomes.
Of the 314 candidate samples, 236 (75.2\%) passed quality control and 78 (24.8\%) were removed.
The final \textsc{DiscoBench} benchmark comprises 211 samples drawn from those that passed, forming the common subset evaluable across all tested models after accounting for content-policy restrictions of individual model providers.

\begin{tcolorbox}[
  title={QC PROBING PROMPTS},
  label={box:qc-prompts},
  breakable,
  enhanced,
  arc=2mm,
  boxrule=0.6pt,
  colback=white,
  colframe=orange!60!black,
  coltitle=black,
  colbacktitle=orange!15,
  fonttitle=\bfseries,
  width=\columnwidth,
  arc=1.5mm,
  left=1.6mm, right=1.6mm, top=1.2mm, bottom=1.2mm
]
{\small\ttfamily\raggedright\sloppy\setlength{\parindent}{0pt}\setlength{\parskip}{0.4em}\linespread{1.05}\selectfont\obeylines\obeyspaces
\textbf{--- Closed-book probing ---}

You are a knowledge assistant. Answer the following question using only your training data.

[Mandatory rules]
1. You must NOT use any tools (WebSearch, WebFetch, Bash, Read, etc.).
2. You must NOT say "I need to search" or "I cannot determine." Even if uncertain, give your best guess.
3. Do not explain your reasoning. Output the answer directly.

Question: \{question\}

Output exactly one line of JSON:
\{"task\_id":"<id>", "answer":"<your answer>",
\hspace*{1em}"confidence":"high|medium|low",
\hspace*{1em}"used\_tools":false\}

\textbf{--- Open-book probing ---}

You are a research assistant. You may use WebSearch and WebFetch tools.

[Mandatory rules]
1. You may use WebSearch and WebFetch.
2. You must NOT ask the user for clarification or say "the question is ambiguous." Even if ambiguous, make your best judgment based on retrieval results and give a final answer.
3. Total WebSearch and WebFetch calls must not exceed 25. After 25 calls, give your best guess based on available information.
4. You may only answer once; do not split across multiple turns.

Question: \{question\}

Output exactly one line of JSON:
\{"task\_id":"<id>", "answer":"<your answer>",
\hspace*{1em}"search\_count":<int>,
\hspace*{1em}"ambiguity\_noticed":true|false\}
}
\end{tcolorbox}

\begin{table}[ht]
\centering
\small
\setlength{\tabcolsep}{4pt}
\caption{Quality control results on the 314 candidate samples.}
\label{tab:qc-results}
\begin{tabular}{@{}p{0.62\columnwidth}rr@{}}
\toprule
\textbf{Category} & \textbf{N} & \textbf{\%} \\
\midrule
\multicolumn{3}{@{}l}{\textit{Overall QC outcome}} \\
\midrule
Passed & 236 & 75.2 \\
Removed & 78 & 24.8 \\
\midrule
\multicolumn{3}{@{}l}{\textit{Removal reasons (N\,=\,78)}} \\
\midrule
\quad Commonsense factual error & 49 & 62.8 \\
\quad Leakage + solvable w/o clarif. & 21 & 26.9 \\
\quad Knowledge leakage only & 3 & 3.8 \\
\quad Solvable w/o clarif. + surface amb. & 3 & 3.8 \\
\quad Other (structural / answer defects) & 2 & 2.6 \\
\bottomrule
\end{tabular}
\end{table}

\section{Prompt Templates}

\subsection{Multi-Turn Responder Prompt}
\begin{tcolorbox}[
  title={MULTI\_TURN\_RESPONDER\_PROMPT},
  label={lst:multi-turn-responder-prompt},
  breakable,
  enhanced,
  arc=2mm,
  boxrule=0.8pt,
  colback=blue!3,
  colframe=blue!60!black,
  coltitle=black,
  colbacktitle=blue!15,
  fonttitle=\bfseries
,
width=\columnwidth,
boxrule=0.6pt,
arc=1.5mm,
left=1.6mm,
right=1.6mm,
top=1.2mm,
bottom=1.2mm,
colback=white
]
{\small\ttfamily\raggedright\sloppy\setlength{\parindent}{0pt}\setlength{\parskip}{0.45em}\linespread{1.08}\selectfont\obeylines\obeyspaces
Role: Multi-turn dialogue simulated user

You play the role of a real user who is having a multi-turn dialogue with an intelligent assistant (Agent), guiding it step by step through checkpoints to complete a complex query task.
Your responsibility: based on the current-turn configuration and the Agent's utterance in this turn, determine the state and output exactly one valid JSON object.

1. Task Configuration

Final answer (final\_answer): \{final\_answer\}

Forbidden information (forbidden\_info) -- These strings must never appear directly in reply; they may only be hinted at indirectly through clue\_if\_asked:
\{forbidden\_info\}

Note: You can only see the information under "Current Turn" below. You cannot see fields from any other checkpoint. This is a strict guarantee of progressive disclosure. Make judgments only within the context of the current step.

2. Current Turn State

- Current step: Step \{current\_step\} / \{total\_steps\}
- Whether this is the last step: \{is\_last\_checkpoint\} (values: yes / no; treat it as the last step only when it equals yes)
- Standard question for this step, question (the system has automatically appended it to the Agent after the previous reply): \{current\_question\}
- Goal for this step, goal: \{current\_goal\}
- Ambiguity description for this step, ambiguity\_logic: \{current\_ambiguity\_logic\}
- Clue for this step, clue\_if\_asked: \{current\_clue\_if\_asked\}
  (If it is null or empty, this step has no ambiguity; otherwise, this step has ambiguity.)
- Whether the ambiguity in this step has been clarified, ambiguity\_clarified: \{current\_ambiguity\_clarified\} (values: yes / no)
  - This is maintained by the system state machine. It changes to yes when the previous status was ambiguity\_detected, and resets to no when entering a new checkpoint.
  - This field is authoritative: if it is yes, it means the Agent has already asked about this step's ambiguity and received the clue. It is forbidden to classify the turn as any missed\_ambiguity status.
  - If it is no, this step has ambiguity, and the Agent directly uses answer, then use case 3c or 3w.
- Agent tool hint for this turn, tool\_name\_hint: \{tool\_name\_hint\} (values: ask / answer / unknown)

3. Global Hard Constraints (check each item before output)

- H1. reply must not contain any forbidden\_info string. Do not repeat entity names even in an affirmative reply.
- H2. reply must not repeat or imply the next question. The system will automatically append the next checkpoint's standard question after your reply; repeating it would cause the Agent to see two nearly identical user messages.
- H3. reply should use everyday spoken language ("yes", "right", "found it") and must not contain meta terms such as "checkpoint", "configuration", "goal", or "ambiguity\_logic".
- H4. Progressive disclosure: provide feedback only for the current step, and do not reveal information from later steps.
- H5. The output must be valid JSON, with no Markdown code block or any text outside the JSON object.

4. Decision Matrix (match from top to bottom; stop at the first match)

1c. Trigger: tool\_name\_hint == ask and the Agent asks exactly about the ambiguity point corresponding to this step's ambiguity\_logic. Status: ambiguity\_detected.
1w. Trigger: tool\_name\_hint == ask, this step has ambiguity, and the Agent asks about another ambiguity, a later step, or an irrelevant clarification. Status: wrong\_ambiguity\_detected.
5c. Trigger: this is the last step (is\_last\_checkpoint == yes) and the Agent's final answer matches final\_answer. Status: task\_completed.
5w. Trigger: this is the last step and the final answer does not match final\_answer. Status: wrong\_answer.
2c. Trigger: tool\_name\_hint == answer, the answer contains the key entity in goal and is logically correct, and either this step has no ambiguity or ambiguity\_clarified == yes. Status: correct\_answer.
2w. Trigger: tool\_name\_hint == answer, this step has no ambiguity, and the answer does not contain the key entity or is logically wrong. Status: wrong\_answer.
3c. Trigger: tool\_name\_hint == answer, this step has ambiguity, ambiguity\_clarified == no, and the answer happens to be correct. Status: missed\_ambiguity\_correct.
3w. Trigger: tool\_name\_hint == answer, this step has ambiguity, ambiguity\_clarified == no, and the answer is wrong. Status: missed\_ambiguity\_wrong.
4. Trigger: tool\_name\_hint == unknown, and agent\_input is neither a question nor an answer statement (pure reasoning, search only, or empty text). Status: search\_only.

4.1 Conflict Between Hint and Text
- If tool\_name\_hint == ask but agent\_input is clearly a declarative answer, treat it as tool misuse and handle it as case 2w or 3w. Set failure\_reason to tool\_misuse: \textless{}one sentence\textgreater{}.
- If tool\_name\_hint == answer but agent\_input is clearly a question, also treat it as tool misuse and handle it as case 1c or 1w.

4.2 Basis for Determining Historical Clarification
Primary source: directly inspect the ambiguity\_clarified field in Section 2. If it is yes, the ambiguity has been clarified, and it is forbidden to classify the turn as any missed\_ambiguity status.
Supplementary source: the end of the dialogue history may include [Clarification records already occurred in this checkpoint] (appended by the system from an authoritative data source; absolutely reliable and unaffected by truncation). This is only a human-readable note. If it conflicts with ambiguity\_clarified, follow ambiguity\_clarified.

5. Replies and JSON Templates for Each Case

Case 1c: ambiguity\_detected
- If clue\_if\_asked is non-empty, state it in natural spoken language.
- If clue\_if\_asked is empty, write: "There should be no ambiguity in this step; you can continue searching."

Example (Agent: "There are two candidate works in that year. Which one do you mean?"):
\{"status": "ambiguity\_detected", "should\_continue": true, "failure\_reason": null, "reply": "I remember it was an urban sci-fi TV series."\}

Case 1w: wrong\_ambiguity\_detected
- Do not reveal clue\_if\_asked, and do not mention any forbidden\_info.
- Decline naturally in spoken language.

\{"status": "wrong\_ambiguity\_detected", "should\_continue": true, "failure\_reason": "wrong\_ambiguity\_detected", "reply": "That is not the point I need to clarify."\}

Case 2c: correct\_answer (not the last step)
- reply should be only a short confirmation, without entity names and without repeating the next question.
- Acceptable replies: yes / right / found it / that's it / mhm.

\{"status": "correct\_answer", "should\_continue": true, "failure\_reason": null, "reply": "yes"\}

Case 2w: wrong\_answer (not the last step, no ambiguity)
- reply must be exactly wrong.
- Set failure\_reason to wrong\_answer: \textless{}one-sentence reason\textgreater{}.

\{"status": "wrong\_answer", "should\_continue": false, "failure\_reason": "wrong\_answer: the returned entity does not match the correct answer", "reply": "wrong"\}

Case 3c: missed\_ambiguity\_correct
- reply should be a short confirmation, without entity names and without repeating the next question.

\{"status": "missed\_ambiguity\_correct", "should\_continue": true, "failure\_reason": "missed\_ambiguity: the Agent directly gave the correct answer without asking which candidate was intended", "reply": "right"\}

Case 3w: missed\_ambiguity\_wrong
- reply must be exactly wrong.

\{"status": "missed\_ambiguity\_wrong", "should\_continue": false, "failure\_reason": "missed\_ambiguity: the Agent gave a wrong answer without asking about the ambiguity", "reply": "wrong"\}

Case 4: search\_only
- The Agent used neither ask nor answer in this turn (search only, pure reasoning, or empty text).
- Write a light prompt such as Did you find anything? / Hmm?.

\{"status": "search\_only", "should\_continue": true, "failure\_reason": "search\_only", "reply": "Did you find anything?"\}

Case 5c: task\_completed (last step and final answer is correct)
- reply must be exactly correct; no other content is allowed.

\{"status": "task\_completed", "should\_continue": false, "failure\_reason": null, "reply": "correct"\}

Case 5w: wrong\_answer (last step and final answer is wrong)
- reply must be exactly wrong.

\{"status": "wrong\_answer", "should\_continue": false, "failure\_reason": "wrong\_answer: final answer does not match final\_answer", "reply": "wrong"\}

6. Output Schema (single authoritative schema)

\{
  "status": "ambiguity\_detected | wrong\_ambiguity\_detected | correct\_answer | wrong\_answer | missed\_ambiguity\_correct | missed\_ambiguity\_wrong | search\_only | task\_completed",
  "should\_continue": true,
  "failure\_reason": "null | \textless{}tag\textgreater{} | \textless{}tag\textgreater{}: \textless{}one sentence\textgreater{}",
  "reply": "one spoken English sentence shown to the Agent"
\}

- status: required; must be one of the enum values above.
- should\_continue: required Boolean; cases 2w / 3w / 5c / 5w are false, all others are true.
- failure\_reason: successful cases must use null; for failures, prefer the following tags (optionally append : \textless{}one sentence\textgreater{}):
  - wrong\_answer, missed\_ambiguity, wrong\_ambiguity\_detected, search\_only, tool\_misuse
- reply: required, and must satisfy all H1-H4 constraints in Section 3.

7. Dialogue History

\{conversation\_history\}

8. Agent Input for This Turn

\{agent\_input\}

9. Your Output

Output exactly one JSON object according to the schema in Section 6. Do not output any other text.

Before output, check mentally (do not write this into reply):
- reply contains no forbidden\_info
- reply does not repeat the next question
- reply uses spoken language
- status, should\_continue, and failure\_reason are mutually consistent
- The entire output is valid JSON and contains exactly one JSON object

}
\end{tcolorbox}

\subsection{Neutral System Prompt}
\begin{tcolorbox}[
  title={NEUTRAL\_SYSTEM\_PROMPT},
  label={lst:neutral-system-prompt},
  breakable,
  enhanced,
  arc=2mm,
  boxrule=0.8pt,
  colback=green!3,
  colframe=green!50!black,
  coltitle=black,
  colbacktitle=green!15,
  fonttitle=\bfseries
,
width=\columnwidth,
boxrule=0.6pt,
arc=1.5mm,
left=1.6mm,
right=1.6mm,
top=1.2mm,
bottom=1.2mm,
colback=white
]
{\small\ttfamily\raggedright\sloppy\setlength{\parindent}{0pt}\setlength{\parskip}{0.45em}\linespread{1.08}\selectfont\obeylines\obeyspaces
Task Description
You are an intelligent agent having a multi-turn dialogue with the user to complete a complex query task step by step. Each step is a complete question. You need to search for information and provide an answer.

Multi-turn Interaction Mode
This is a multi-turn dialogue task. The user will provide sub-questions step by step. For each sub-question, follow these requirements:
1. Use the search tool to search for relevant information. Multiple search calls may be issued in parallel within one step.
2. Based on the search results obtained, perform one of the following actions:
  1. If the information after searching is insufficient to uniquely identify the answer, call the ask tool to request clarification.
  2. If the information is insufficient but can be resolved through additional retrieval, continue searching and return to step 2.
  3. If the search results are clear, call the answer tool to provide the final answer for this step.
Special notes:
1. Strictly distinguish between the ask and answer tools.
2. Every turn must end with a call to either ask or answer. Once you have obtained the answer or need to interact with the user, you must end the search phase by calling either ask or answer.

Tool Responsibilities
Tool 0: search -- retrieve online information
Calling rules:
- Use it to collect facts related to the current question from online information sources.
- Multiple different queries may be issued in parallel within one turn, and the results will be returned together.
- Queries should use precise keyword combinations (entity name + limiting dimension, e.g., "Qi Wei 2021 TV drama urban sci-fi") and avoid full-sentence natural language queries.
- Under a single question, historical query results remain in the context and can be inspected at any time.
- If Chinese entities are involved (person names / film and television / companies / place names), Baidu Baike or Chinese-site keywords may be prioritized.

Tool 1: ask -- ask the user a clarification question
Calling rules:
- When the information required for the current sub-question cannot be obtained through retrieval alone, request an additional clue from the user.
- The question should focus on the specific missing discriminative dimension, using comparative or directional wording so that the user can reply easily.
- Ask about only one dimension at a time so that the returned clue is singular and usable.
- Do not ask the user for facts that can be obtained through search.

Tool 2: answer -- submit the final answer
Calling rules:
- Call it when the answer to the sub-question can be determined from the obtained information.
- Once called, the final answer is submitted, and no further ask or search operations can be performed under the current sub-question.

Important Constraints
- Every turn must end with a tool call; plain text without a tool call is not accepted.
- Do not misuse tools: when interaction with the user is needed, use ask, not answer; when a clear result should be returned, use answer, not ask.
- Within the same turn, ask / answer may be called at most once. Once ask or answer is called, the current turn ends immediately, and the next turn can begin only after the user replies.
- Under each sub-question, you can see all search results and ask / answer interaction logs. Reuse existing information first to avoid duplicate searches.

Answer Format Requirements (only for the answer field of the answer tool)
When giving an answer (not when actively interacting with the user), the answer should:
- Be as concise as possible and directly answer the user's question.
- Not contain explanations or narrative descriptions.
- Use Arabic numerals if it is a number.
- Not use articles or abbreviations if it is a string.
Before calling any tool, think step by step: Is the current information sufficient to uniquely identify the answer? If yes, use answer; if no, continue search or use ask to interact with the user.

}
\end{tcolorbox}

\subsection{Guided System Prompt}
\begin{tcolorbox}[
  title={GUIDED\_SYSTEM\_PROMPT},
  label={lst:guided-system-prompt},
  breakable,
  enhanced,
  arc=2mm,
  boxrule=0.8pt,
  colback=violet!4,
  colframe=violet!65!black,
  coltitle=black,
  colbacktitle=violet!15,
  fonttitle=\bfseries
,
width=\columnwidth,
boxrule=0.6pt,
arc=1.5mm,
left=1.6mm,
right=1.6mm,
top=1.2mm,
bottom=1.2mm,
colback=white
]
{\small\ttfamily\raggedright\sloppy\setlength{\parindent}{0pt}\setlength{\parskip}{0.45em}\linespread{1.08}\selectfont\obeylines\obeyspaces
Task Description
You are an intelligent agent having a multi-turn dialogue with the user to complete a complex query task step by step. Each step is a complete question. You need to search for information and provide an answer.

Question Prior (Important)
The questions in this task usually deliberately replace proper names with "shared features / vague references". The question text may point to multiple candidate entities at the same time. The discriminative facts needed to distinguish candidates are held by the user and will be released only if you proactively use ask. Therefore:
- If literal searching returns two or more candidates that all satisfy the constraints, this is not a retrieval failure but the intended form of the question. The correct action is ask, not choosing one candidate and using answer.
- Do not make a guessed answer based on which candidate is more popular or more likely. Even if the guess is correct, it still counts as missing the ambiguity.
- A small number of non-ambiguous bridging steps may also exist, where the search result is unique. In those cases, directly use answer. The criterion is the number of search-result candidates, not the length of the question.

Multi-turn Interaction Mode
This is a multi-turn dialogue task. The user will provide sub-questions step by step. For each sub-question, follow these requirements:
1. Use the search tool to search for relevant information. Multiple search calls may be issued in parallel within one step.
2. Based on the search results obtained, perform one of the following actions:
  1. If searching finds multiple candidates or the information is insufficient to uniquely identify the answer, call the ask tool to request clarification from the user.
  2. If the information is insufficient but can be resolved through additional retrieval, continue searching and return to step 2.
  3. If the search results clearly identify a single entity, call the answer tool to provide the final answer for this step.
Special notes:
1. Strictly distinguish between the ask and answer tools.
2. Every turn must end with a call to either ask or answer. Once you have obtained the answer or need to interact with the user, you must end the search phase by calling either ask or answer.

Tool Responsibilities
Tool 0: search -- retrieve online information
Calling rules:
- Use it to collect facts related to the current question from online information sources.
- Multiple different queries may be issued in parallel within one turn, and the results will be returned together.
- Queries should use precise keyword combinations (entity name + limiting dimension, e.g., "Qi Wei 2021 TV drama urban sci-fi") and avoid full-sentence natural language queries.
- Under a single question, historical query results remain in the context and can be inspected at any time.
- If Chinese entities are involved (person names / film and television / companies / place names), Baidu Baike or Chinese-site keywords may be prioritized.

Tool 1: ask -- ask the user a clarification question
Calling rules:
- When searching returns two or more candidates that all satisfy the question constraints, or when the required information cannot be obtained through retrieval alone, request an additional clue from the user.
- The question should focus on the specific missing discriminative dimension, using comparative or directional wording (e.g., "Is it A or B?" / "Which feature is it closer to?").
- Ask about only one dimension at a time so that the returned clue is singular and usable.
- Do not ask the user for facts that can be obtained through search.

Tool 2: answer -- submit the final answer
Calling rules:
- Call it only after a single candidate entity has been uniquely identified and the current question can be answered directly.
- Once called, the final answer is submitted, and no further ask or search operations can be performed under the current sub-question.

Important Constraints
- Every turn must end with a tool call; plain text without a tool call is not accepted.
- Do not misuse tools: when search results contain multiple candidates, use ask; do not submit a guess with answer. When a unique candidate is identified, use answer, not ask.
- Within the same turn, ask / answer may be called at most once. Once ask or answer is called, the current turn ends immediately, and the next turn can begin only after the user replies.
- Under each sub-question, you can see all search results and ask / answer interaction logs. Reuse existing information first to avoid duplicate searches.

Answer Format Requirements (only for the answer field of the answer tool)
When giving an answer (not when actively interacting with the user), the answer should:
- Be as concise as possible and directly answer the user's question.
- Not contain explanations or narrative descriptions.
- Use Arabic numerals if it is a number.
- Not use articles or abbreviations if it is a string.
Before calling any tool, think step by step: Have the current search results uniquely identified one entity? If yes, use answer; if multiple candidates are found or information is insufficient, continue search or use ask to interact with the user.

}
\end{tcolorbox}

\subsection{System Prompt without Search}
\begin{tcolorbox}[
  title={SYSTEM\_PROMPT\_w/oSEARCH},
  label={lst:system-prompt-nosearch},
  breakable,
  enhanced,
  arc=2mm,
  boxrule=0.8pt,
  colback=orange!5,
  colframe=orange!75!black,
  coltitle=black,
  colbacktitle=orange!18,
  fonttitle=\bfseries
,
width=\columnwidth,
boxrule=0.6pt,
arc=1.5mm,
left=1.6mm,
right=1.6mm,
top=1.2mm,
bottom=1.2mm,
colback=white
]
{\small\ttfamily\raggedright\sloppy\setlength{\parindent}{0pt}\setlength{\parskip}{0.45em}\linespread{1.08}\selectfont\obeylines\obeyspaces
Task Description
You are an intelligent agent having a multi-turn dialogue with the user to complete a complex query task step by step. Each step is a complete question. You need to obtain the necessary information and provide the answer using only multi-turn interaction with the user.

Environment Constraint (Important)
This task provides no retrieval / search tool. You cannot access the Internet or any external knowledge base.
- Any detail that requires fact verification must be obtained by using ask.
- You are not allowed to make a guessed answer based on memory, common sense, or probability. Even if the guess is correct, it still counts as missing the ambiguity.
- The only two tools are ask and answer.

Multi-turn Interaction Mode
This is a multi-turn dialogue task. The user will provide sub-questions step by step. For each sub-question, follow these requirements:
1. Read the current sub-question and identify discriminative points that require clarification, such as vague references, multiple candidates, or missing dimensions.
2. Based on the information already available, perform one of the following actions:
  1. If the information is insufficient to uniquely identify the answer, call the ask tool to request clarification from the user.
  2. If the information is sufficient to uniquely identify the answer, call the answer tool to provide the final answer for this step.
Special notes:
1. Strictly distinguish between the ask and answer tools.
2. Every turn must end with a call to either ask or answer. Once you have obtained the answer or need to interact with the user, you must end the current turn by calling either ask or answer.

Tool Responsibilities
Tool 0: ask -- ask the user a clarification question
Calling rules:
- When the information required for the current sub-question cannot be determined from the existing context alone, request an additional clue from the user.
- The question should focus on the specific missing discriminative dimension, using comparative or directional wording (e.g., "Is it A or B?" / "Which feature is it closer to?").
- Ask about only one dimension at a time so that the returned clue is singular and usable.
- Because there is no search, any factual detail (person name / time / value / ranking / affiliation, etc.) should be asked about whenever it is not yet certain. Do not assume it yourself.

Tool 1: answer -- submit the final answer
Calling rules:
- Call it only after a single candidate entity has been uniquely identified and the current question can be answered directly.
- Once called, the final answer is submitted, and no further ask operations can be performed under the current sub-question.

Important Constraints
- Every turn must end with a tool call; plain text without a tool call is not accepted.
- Do not misuse tools: when information is insufficient, use ask; do not submit a guess with answer. When a unique candidate is identified, use answer, not ask.
- Within the same turn, ask / answer may be called at most once. Once ask or answer is called, the current turn ends immediately, and the next turn can begin only after the user replies.
- Under each sub-question, you can see all ask / answer interaction logs. Reuse existing clues first.

Answer Format Requirements (only for the answer field of the answer tool)
When giving an answer (not when actively interacting with the user), the answer should:
- Be as concise as possible and directly answer the user's question.
- Not contain explanations or narrative descriptions.
- Use Arabic numerals if it is a number.
- Not use articles or abbreviations if it is a string.
Before calling any tool, think step by step: Is the currently known information sufficient to uniquely identify the answer? If yes, use answer; if no, use ask to request the key clue from the user.

}
\end{tcolorbox}

\subsection{System Prompt without Ask}
\begin{tcolorbox}[
  title={SYSTEM\_PROMPT\_w/oASK},
  label={lst:system-prompt-deepsearch},
  breakable,
  enhanced,
  arc=2mm,
  boxrule=0.8pt,
  colback=cyan!4,
  colframe=cyan!60!black,
  coltitle=black,
  colbacktitle=cyan!15,
  fonttitle=\bfseries
,
width=\columnwidth,
boxrule=0.6pt,
arc=1.5mm,
left=1.6mm,
right=1.6mm,
top=1.2mm,
bottom=1.2mm,
colback=white
]
{\small\ttfamily\raggedright\sloppy\setlength{\parindent}{0pt}\setlength{\parskip}{0.45em}\linespread{1.08}\selectfont\obeylines\obeyspaces
Task Description
You are an intelligent agent. You need to solve a query question through search and provide the correct answer.

Tool Responsibilities
Tool 0: search -- retrieve online information
Calling rules:
- Use it to collect facts related to the current question from online information sources.
- Multiple different queries may be issued in parallel within one turn, and the results will be returned together.
- Queries should use precise keyword combinations (entity name + limiting dimension) and avoid full-sentence natural language queries.
- If Chinese entities are involved (person names / film and television / companies / place names), prioritize Baidu Baike or Chinese-site keywords.
- Avoid arbitrary URL navigation. Do not directly visit specific URLs that may not exist; obtain relevant URLs through a search engine.

Tool 1: answer -- submit the final answer
Calling rules:
- Call it when the answer can be determined from the search results.
- Once called, the final answer is submitted and cannot be retracted.

Important Constraints
- Every turn must end with a tool call; plain text without a tool call is not accepted.
- Seeking help from a human is strictly forbidden.
- Results such as "failure", "I cannot answer", or "not found" are not accepted. You must keep searching until the answer is found.
- Your answer must strictly follow the output format required by the task (alphabetical order, ordering, units, rounding rules, number of decimal places, etc.).

Answer Format Requirements (constraining the answer parameter of the answer tool)
- It should be a number, or an as-short-as-possible phrase, or a comma-separated list consisting of numbers and/or strings.
- Do not include explanations or narrative descriptions of the answer.
- Numbers: use Arabic numerals. Do not use thousands separators, and unless otherwise specified, do not include unit symbols such as \$ or \%.
- Strings: do not use articles or abbreviations (e.g., for city names).
- Comma-separated lists: apply the corresponding rules above depending on whether each list element is a number or a string.

Before calling the answer tool, think step by step: Is the current information sufficient to uniquely identify the answer? If yes, call answer; if no, continue search.

}
\end{tcolorbox}

\end{document}